\documentclass[10pt,twocolumn,letterpaper]{article}

\usepackage{wacv}              

\usepackage{graphicx}
\usepackage{amsmath}
\usepackage{amssymb}
\usepackage{booktabs}

\usepackage{pifont}
\newcommand{\cmark}{\ding{51}}%

\usepackage{multirow}
\usepackage[accsupp]{axessibility}  

\usepackage[pagebackref,breaklinks,colorlinks]{hyperref}

\usepackage[capitalize]{cleveref}
\crefname{section}{Sec.}{Secs.}
\Crefname{section}{Section}{Sections}
\Crefname{table}{Table}{Tables}
\crefname{table}{Tab.}{Tabs.}

\def\wacvPaperID{993} 
\def\confName{WACV}
\def\confYear{2024}

\begin{document}

\title{S$^3$AD: Semi-supervised Small Apple Detection in Orchard Environments}

\author{Robert Johanson$^\ast$,
Christian Wilms$^\ast$, 
Ole Johannsen, and
Simone Frintrop \\
Computer Vision Group, University of Hamburg, Germany \\
$^\ast$ denotes equal contribution\\
{\tt\small \{firstname\}.\{lastname\}@uni-hamburg.de}}

\maketitle

\begin{abstract}

Crop detection is integral for precision agriculture applications such as automated yield estimation or fruit picking. However, crop detection, e.g., apple detection in orchard environments remains challenging due to a lack of large-scale datasets and the small relative size of the crops in the image. In this work, we address these challenges by reformulating the apple detection task in a semi-supervised manner. To this end, we provide the large, high-resolution dataset MAD\footnote{MAD is available at \url{www.inf.uni-hamburg.de/mad}} comprising 105 labeled images with 14,667 annotated apple instances and 4,440 unlabeled images. Utilizing this dataset, we also propose a novel \textbf{S}emi-\textbf{S}upervised \textbf{S}mall \textbf{A}pple \textbf{D}etection system S$^3$AD\footnote{Code is available at \url{www.github.com/RbtJhs/SSSAD}} based on contextual attention and selective tiling to improve the challenging detection of small apples, while limiting the computational overhead. We conduct an extensive evaluation on MAD and the MSU dataset, showing that S$^3$AD substantially outperforms strong fully-supervised baselines, including several small object detection systems, by up to $14.9\%$. Additionally, we exploit the detailed annotations of our dataset w.r.t. apple properties to analyze the influence of relative size or level of occlusion on the results of various systems, quantifying current challenges.
\end{abstract}

\section{Introduction}
\label{sec:intro}

Pre-harvest yield estimation is an integral part of agriculture for efficiently planning harvest, sales, transportation, and storage of crops~\cite{wang2013automated,hani2018apple,akiva2020finding,stein2016image}. Yield estimation usually relies on labor-intensive manual counting in sample locations~\cite{wang2013automated,hani2018apple,akiva2020finding,stein2016image}, as well as weather information and historical data~\cite{wang2013automated,hani2018apple,akiva2020finding}. However, such estimations are inaccurate due to natural variances in fruit load, soil,- and light exposure~\cite{koirala2019deep,stein2016image}. Recently, precision agriculture has received significant attention, leading to vision-based systems for automated yield estimation~\cite{hani2018apple,akiva2020finding}, quality control~\cite{tian2019apple}, or fruit picking~\cite{hua2019recent}.

\begin{figure}
\centering
\begin{tabular}{cc}
\includegraphics[width = .45\linewidth]{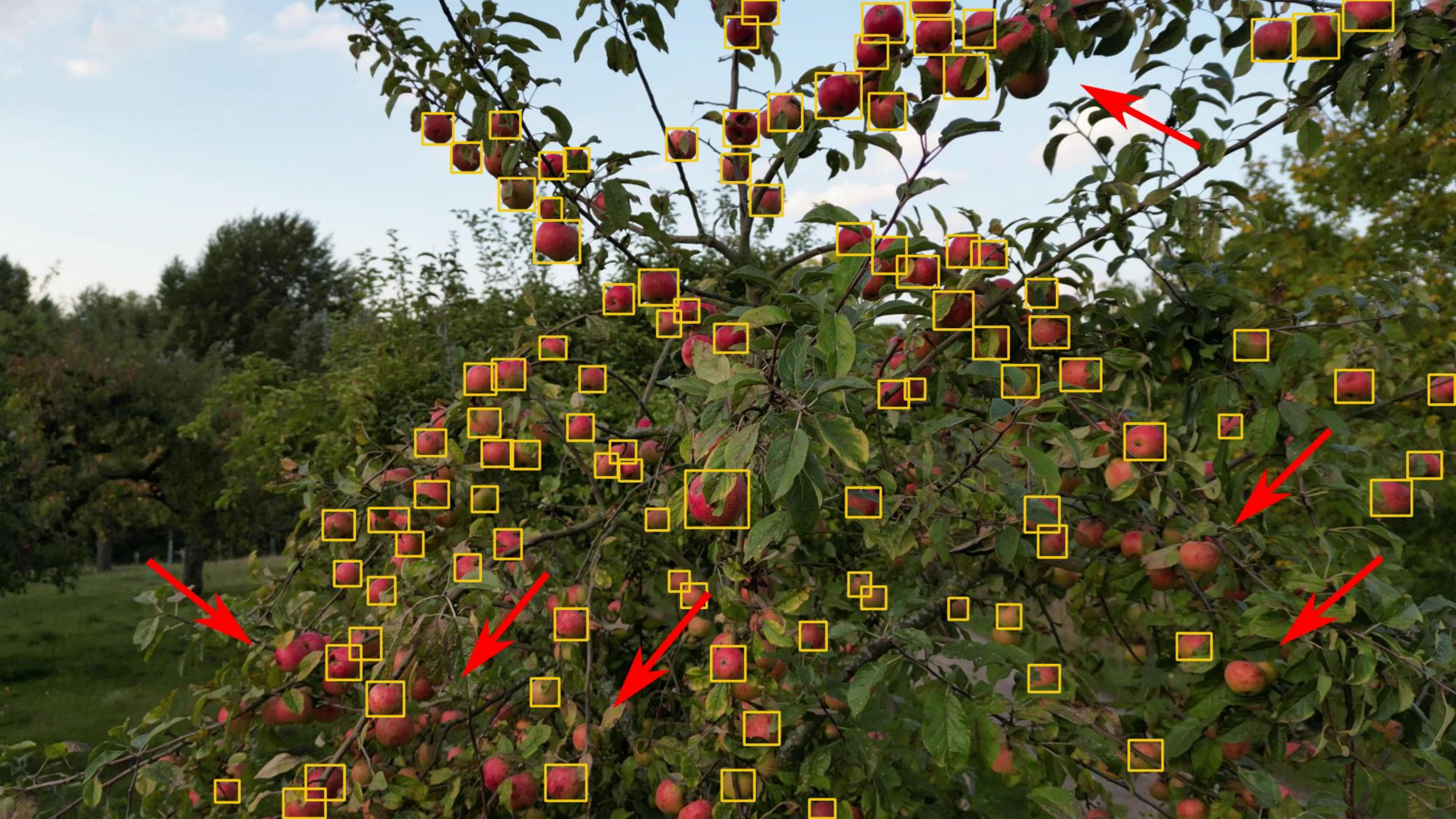} &
\includegraphics[width = .45\linewidth]{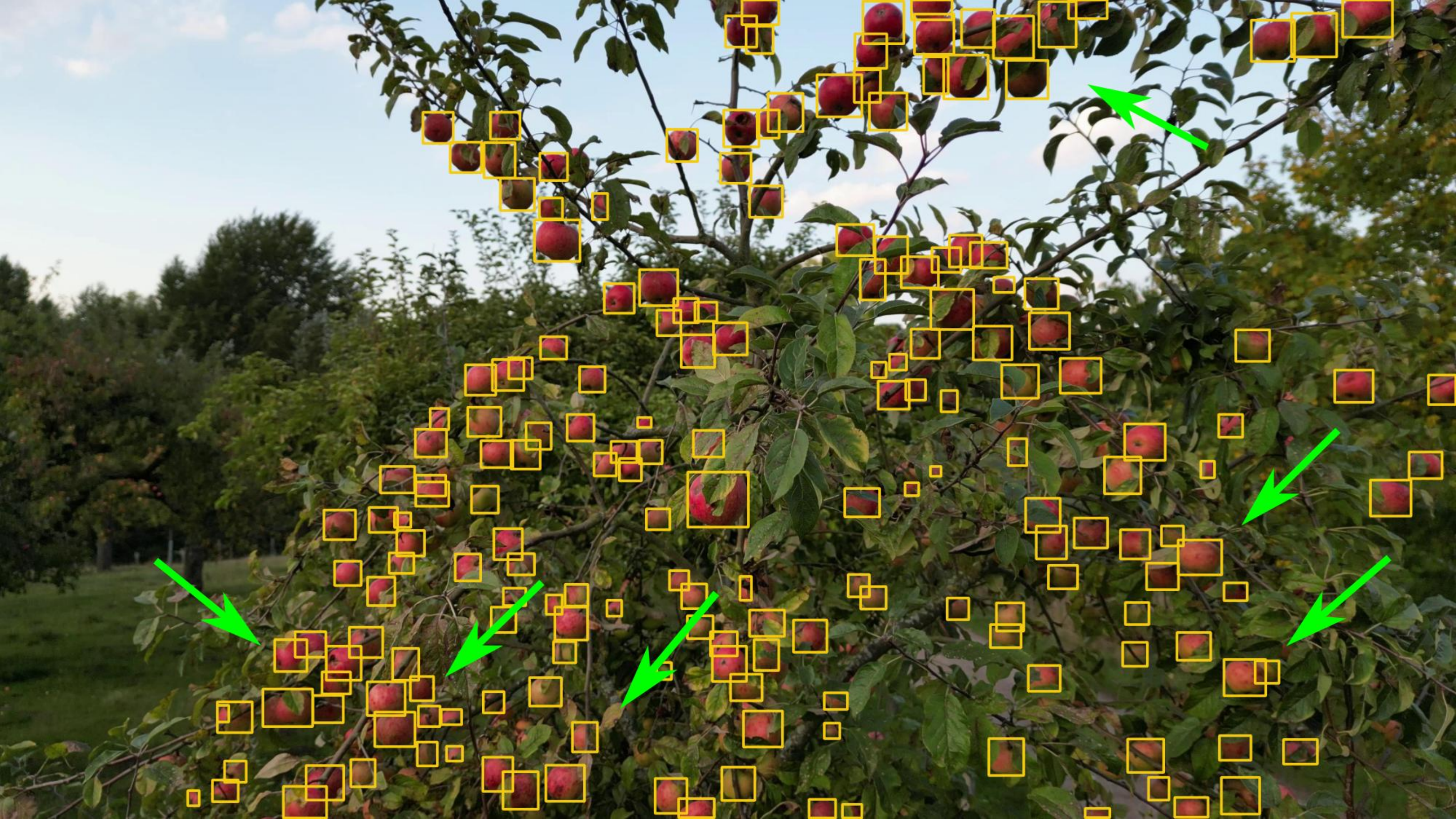}\\

{\footnotesize Faster R-CNN+FPN}&{\footnotesize S$^3$AD (ours)}
\end{tabular}
\caption{Apple detection results using Faster R-CNN+FPN~\cite{ren2015faster,lin2017feature} and our proposed, semi-supervised small apple detection system S$^3$AD with selective tiling on a test image of our MAD dataset. Red arrows denote missed apples, while green arrows denote the detection of previously missed apples by the other system.
}
\label{fig:cover_fig}
\end{figure}

An essential task in the development of such systems is the reliable detection of crops~\cite{hani2020minneapple}. Despite recent advancements~\cite{hani2020minneapple,hani2018apple,tian2019apple,stein2016image,akiva2020finding}, this task remains challenging. For instance, the detection of apples in orchards remains difficult due to the complex environment caused by multiple factors: (i) the dense apple distribution, (ii) occlusion and shadow induced by other crops or leaves, and (iii) the small size of the apples compared to the trees. These effects are also visible from the example in Fig.~\ref{fig:cover_fig} depicting a typical apple tree in an orchard environment. Especially challenging is the small apple size, since the performance of object detectors decreases significantly for small objects~\cite{chen2020survey}, due to the inherent down-sampling in CNNs and the limited GPU resources preventing the processing of entire high-resolution images. Moreover, the limited availability of data~\cite{bargoti2017deep,hani2020minneapple} is another challenge in apple detection. 

To address the aforementioned challenges, several approaches were proposed in general object detection. To cope with a limited amount of data, transfer learning~\cite{tan2018survey} is used to utilize features learned on tasks with less annotation effort. Furthermore, semi-supervised systems were proposed to incorporate unlabeled data in object detection by consistency~\cite{jeong2019consistency,tang2021proposal} and pseudo-label methods~\cite{sohn2020simple,xu2021end}. To improve the challenging detection of small objects, several directions were explored~\cite{cheng2023towards} including multiscale feature extraction~\cite{lin2017feature,liu2018path,singh2018sniper,yang2022querydet} to improve the feature representation of small objects by learning scale-specific features, attention-based methods~\cite{yang2019scrdet,lu2021attention,yang2022scrdet++} that aim to select relevant areas or features within the network, and tiling strategies~\cite{ozge2019power,yang2019clustered,wilms2021airline} that improve the relative size of objects on input level. 


In this paper, we address the challenging detection of apples in orchard environments by reformulating the problem in a semi-supervised manner. To tackle this problem, we propose a novel semi-supervised small apple detection system S$^3$AD focusing on small apples and a new large-scale apple detection dataset MAD for semi-supervised learning. MAD consists of 4,545 high-resolution images from apple orchards with 14,667 manually annotated apples across 105 images. The remaining images support semi-supervised learning. To address the problem of apple detection in a semi-supervised way and improving the challenging detection of small apples, we propose S$^3$AD. It is composed of three main modules: (i) an object detector trained with the semi-supervised pseudo-labeling framework Soft Teacher~\cite{xu2021end}, which allows us to utilize the large amount of unlabeled data in our dataset, (ii) the TreeAttention module that leverages the contextual relationship between the apples and the tree crowns to localize regions of interest, and (iii) a selective tiling module that crops tiles from the regions of interest, enabling the object detector to utilize the full image resolution increasing the detection performance on small apples. In our extensive evaluation on two datasets, we show the strong results of S$^3$AD on small apples and apples across all sizes, utilizing contextual attention and selective tiling.

In summary, the contributions of this paper are threefold:   
\begin{itemize}
  \item  We reformulate apple detection as a semi-supervised task limiting the labeling effort and publish a dataset, MAD, containing 105 labeled and 4,440 unlabeled high resolution-images with 14,667 manually annotated apples that facilitates the new formulation.
  \item We propose a novel apple detection system, S$^3$AD, that utilizes semi-supervised learning, contextual attention, and selective tiling to address the limited amount of labeled data and the small apple size. 
  \item We validate our approach with an extensive evaluation of S$^3$AD on MAD and the MSU dataset, comparing it to strong fully-supervised small object detection systems and assessing the effect of three apple properties.
\end{itemize}

\section{Related Work}
\label{sec:relWork}
This section briefly reviews related works on crop detection, crop detection datasets, and small object detection.
\paragraph{Crop Detection} The detection of crops in precision agriculture is mainly based on variations of standard object detectors. For instance, \cite{lawal2021tomato} modify YOLOv3 for detection of the tomatoes, while \cite{yuan2020robust} adapt SSD. \cite{mu2020intact} utilize Faster R-CNN and add image stitching as well as tiling steps to process rows of plants. For mango detection, \cite{koirala2019deep}~modify YOLOv2 and \cite{roy2022real} propose a modified YOLOv4. 

For apple detection, \cite{bargoti2017deep} and \cite{CHU2023100284} employ Faster R-CNN with standard tiling and an occlusion-aware detection module, respectively. Moving to YOLO, \cite{tian2019apple}~propose a YOLOv3 variation with a DenseNet backbone to detect apples in different growth stages. In~\cite{kuznetsova2020using}, the performance of YOLOv3 is enhanced by pre- and post-processing steps. Recently, the authors of~\cite{jiang2022fusion} enhanced YOLOv4 with non-local feature-level attention, and a convolution block attention module to detect apples in low-resolution images.

In contrast, we address apple detection in a semi-supervised manner, and focus on detecting small apples by introducing contextual attention and selective tiling.

\paragraph{Crop Detection Datasets}
Datasets for crop detection exist for various crops~\cite{lu2020survey}. However, most crop detection datasets are limited in size. According to~\cite{hani2020minneapple}, the largest apple detection dataset contains 1,404 labeled images~\cite{stein2016image} with 7,065 annotated instances on low-resolution images. This also applies to the apple detection dataset by~\cite{bargoti2017deep}, containing 841 images with 5,765 labeled instances. The MinneApple dataset~\cite{hani2020minneapple} contains more labeled instances, with 41,325 labeled instances in 1,001 images of moderate resolution~($1280\times720$). Recently, the MSU Apple Dataset V2~\cite{CHU2023100284} containing 14,518 annotated apples in 900 closeup images of apple tree crowns was proposed.

In contrast to these datasets, our proposed dataset MAD is well-suited for semi-supervised apple detection containing labeled and unlabeled images. Moreover, the dataset is larger than existing ones with 4,545 high-resolution images.

\paragraph{Small Object Detection}
\label{sec:small_object_detection}
The problem of small object detection has been addressed with different strategies. See~\cite{cheng2023towards} for an extensive survey. A popular strategy is the use of multiscale or scale-specific features to improve the representation of small objects. While\cite{lin2017feature} learn scale-specific features in a bottom-up and top-down manner, \cite{liu2018path}~ recombine these features for an improved multiscale representation. \cite{singh2018analysis}~and \cite{singh2018sniper} improve the training strategy of object detectors by reducing the intra-scale noise and move from a feature pyramid to an image pyramid. \cite{najibi2019autofocus}~improve the approach's efficiency with a coarse-to-fine detection. Improving the expressiveness of scale-specific features, \cite{li2019scale}~adapt the receptive field for different scales of objects. Recently, \cite{yang2022querydet}~ employed a coarse-to-fine query-based detection mechanism on successively higher-resolution feature maps.

Besides improved feature representation, several works utilize tiling~\cite{ozge2019power,yang2019clustered,wilms2021airline} or super-resolution~\cite{li2017perceptual,bai2018sod,noh2019better} to improve the spatial resolution of features. Another line of work uses attention mechanisms to highlight features or locations of small objects~\cite{yang2019scrdet,lu2021attention,yang2022scrdet++}. Our apple detection system S$^3$AD is most related to the multiscale and tiling approaches. However, it is explicitly designed to address the detection of apples in a semi-supervised framework, e.g., utilizing domain knowledge to learn contextual attention.




\section{Data Acquisition \& Dataset}
\label{sec:data_acqui}
To facilitate our novel semi-supervised formulation of the apple detection task, we propose the Monastery Apple Dataset (MAD). The data was acquired in collaboration with a monastery in Bad Oldesloe, Germany. We collected video data from 16 trees of the monastery's apple orchard with a DJI-Mini 3 Pro drone that has a resolution of 4k~($3840\times 2160$). The data collection was conducted over the course of one month in September 2022 and during different lighting conditions to ensure diversity.

\label{sec:dataset}
\begin{table}[t]
\centering
\caption{Statistics of MAD. The columns show the number of images and the number of labeled instances per dataset split.}
\label{tab:dataset_stats}
\begin{tabular}{lcccc}
\toprule
Split  & Images  &  Labeled Instances\\ \hline

train &  66 & 10089\\
val &  12 & 1288\\
test &  27 & 3290\\
unlabeled &  4440 & n/a\\
\bottomrule
\end{tabular}
\end{table}

\begin{figure}
\centering
\begin{tabular}{cc}
\includegraphics[width = .45\linewidth]{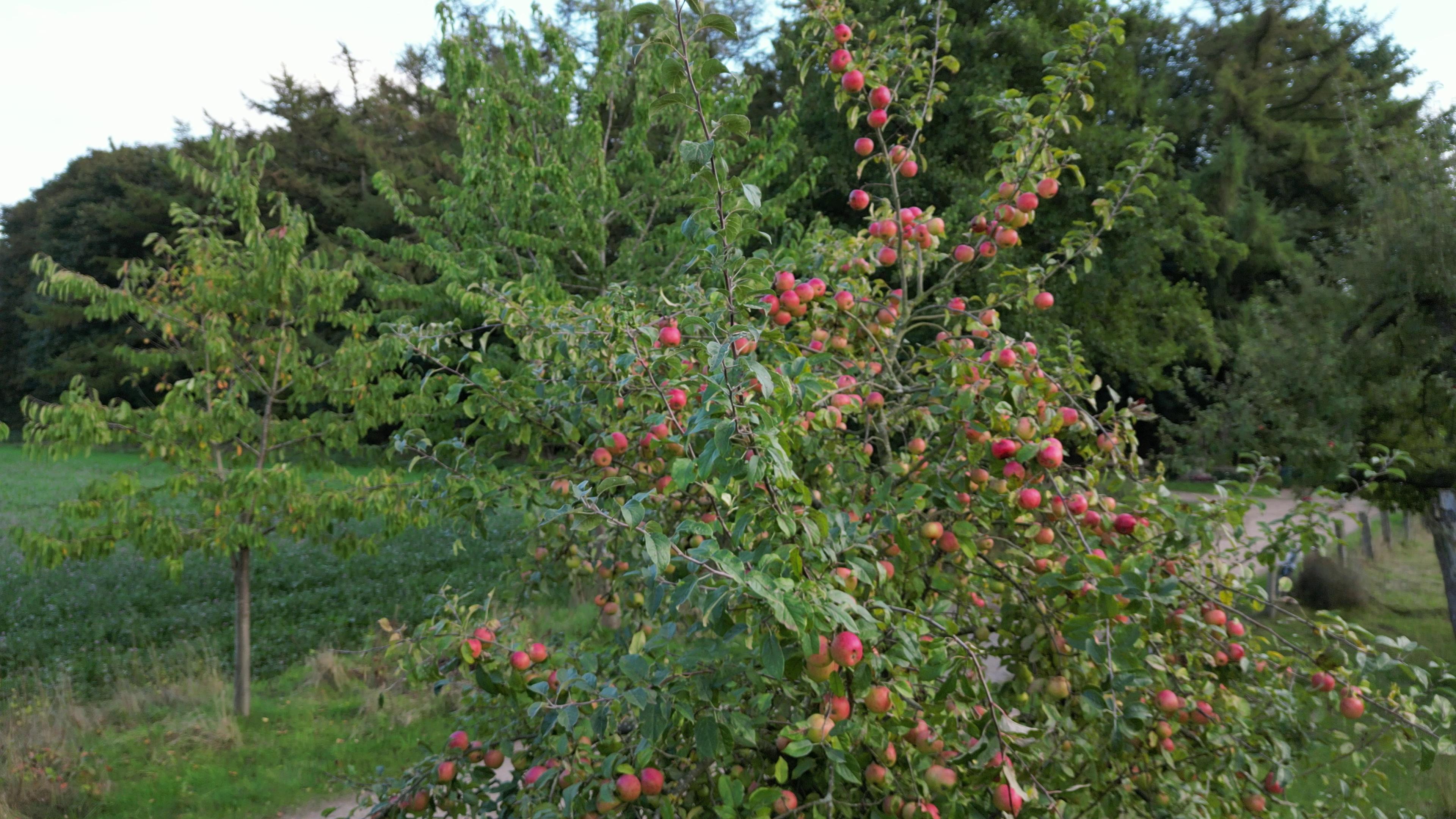} &
\includegraphics[width = .45\linewidth]{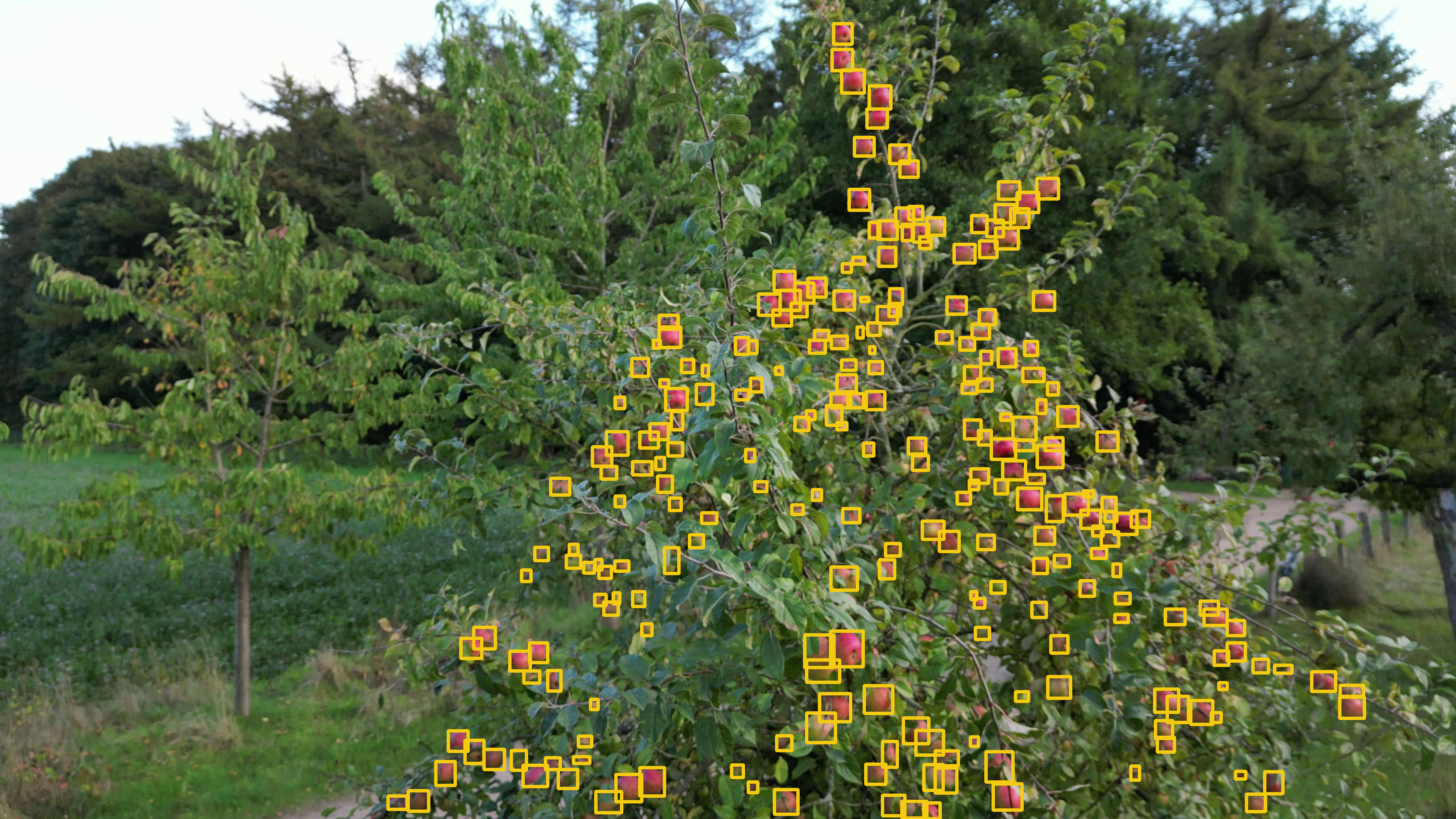}\\
\end{tabular}
\caption{An example image (left) and the corresponding ground truth (right) from the test split of our dataset MAD.}
\label{fig:dataset_example}
\end{figure}

To generate the training, validation, and test splits, we first assigned the videos of 6~(training), 2~(validation), and 4~(test) trees to the respective splits. From the videos, we manually selected images for annotation to maximize data diversity. The training, validation, and test splits consist of 66, 12, and 27 frames with 10,089, 1,288, and 3,290 annotated instances as shown in~\cref{tab:dataset_stats}. The apples were manually annotated with bounding boxes. An example image with ground truth from our dataset is shown in Fig. \ref{fig:dataset_example}. Moreover, we automatically assigned three properties to each annotated apple representing the relative size, the level of occlusion, and the lighting conditions~(see supp. mat. for details and statistics). This enables us, to evaluate the influence of such conditions on our system. The 4,440 unlabeled images in the training split are composed of frames from the 6 training trees that were not annotated and frames from the four remaining videos/trees that were not assigned to the other splits.


\section{Method}
\label{sec:method}
\begin{figure*}[h]
	 \centerline{\includegraphics[width=1.0\textwidth]{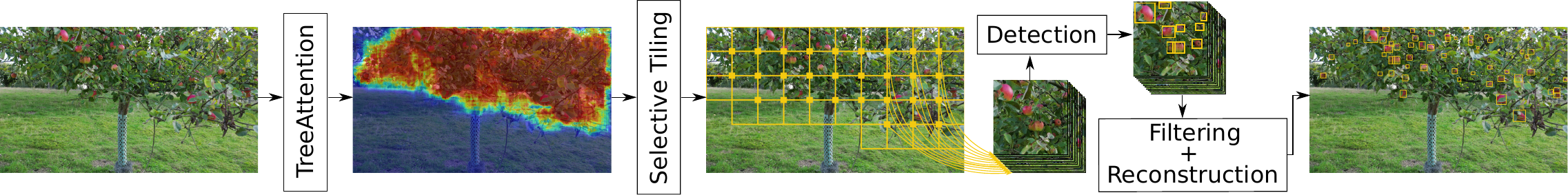}}
	 {\caption{System figure of our proposed semi-supervised small apple detection approach S$^3$AD. First, an attention map is generated with our TreeAttention module. The attention map is used by our selective tiling module to crop the most promising image regions into a set of overlapping tiles, which are subsequently processed by a semi-supervised Faster R-CNN object detector. During filtering and reconstruction, the per-tile results are merged. }\label{fig:pipeline}}
\end{figure*}

This section introduces our novel \textbf{S}emi-\textbf{S}upervised \textbf{S}mall \textbf{A}pple \textbf{D}etection system S$^3$AD that specifically addresses the challenges of small relative apple size and a limited amount of annotated data. S$^3$AD consists of four steps, as shown in~\cref{fig:pipeline}. Given an input image, our new TreeAttention module~(see~\cref{sec:tree_attention}) utilizes contextual information to highlight the most promising image regions for apple detection. These regions are subsequently extracted by our selective tiling module, explained in~\cref{sec:tiling}, and cropped into a set of overlapping tiles. Tiling high-resolution images is advantageous since it increases the relative size of the apples and allows the object detector to process the image without initial down-sampling. Next, we train a semi-supervised Faster R-CNN object detector~\cite{ren2015faster} with FPN backbone~\cite{lin2017feature} in the Soft Teacher framework~(see~\cref{sec:detection})~\cite{xu2021end}. The FPN backbone improves the detection of small apples, while the semi-supervised training in the Soft Teacher framework utilizes the unlabeled data. Finally, we merge the per-tile results in a filtering and reconstruction step, detailed in~\cref{sec:filtering}. 


\subsection{Tree Attention}
\label{sec:tree_attention}
In apple detection, we seek to detect the apples in the tree crown. Hence, given an image, S$^3$AD only needs to search the area of the tree crown, which substantially reduces the search space. Generally, this is known as contextual information and frequently used in small object detection~\cite{chen2020survey}. Therefore, we propose TreeAttention as contextual attention module in S$^3$AD to focus processing on the most prominent tree crowns in the input image. We will use this attention in~\cref{sec:tiling}, to guide S$^3$AD's selective tiling.



\begin{figure}[t]
        \centering
        \begin{subfigure}[b]{0.49\linewidth}
            \centering
            \includegraphics[width=\linewidth]{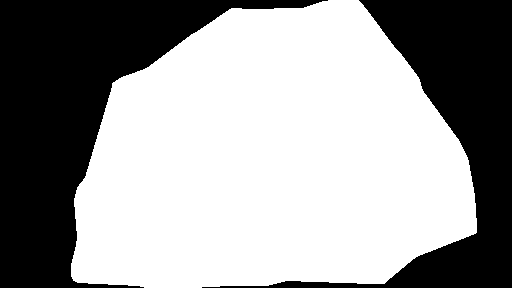}
            \caption{Filled Alpha Shape Contour}
            \label{fig:alpha}
            
        \end{subfigure}
        \hfill
        \begin{subfigure}[b]{0.49\linewidth}
            \centering
            \includegraphics[width=\linewidth]{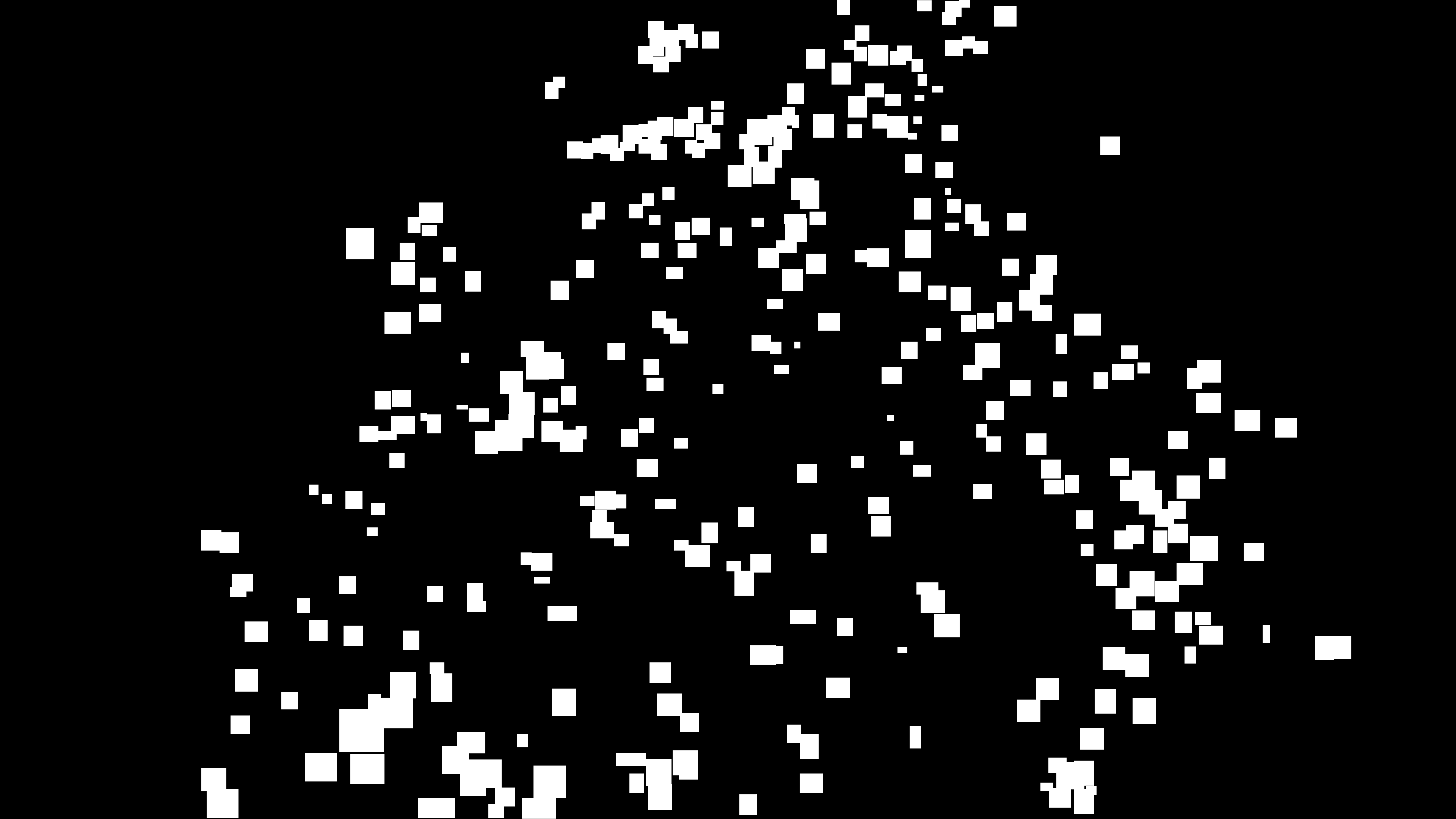}
            
            \caption{Bounding Boxes}
            \label{fig:boxes}
        \end{subfigure}
        \hfill
        \vspace{-5pt}
        \caption{Example of the filled alpha shape (a) that is used to train TreeAttention and the corresponding bounding boxes (b).}
        \label{fig:alpha_shape_vis}
    \end{figure}

Our TreeAttention module is based on a simple residual U-Net architecture \cite{ronneberger2015u} comprising three stages with one residual block each in the encoder and decoder, respectively. The final result of our TreeAttention module is an attention map with values between 1~(high attention) and 0~(low attention) indicating the rough locations of apples. Since no precise localization is necessary in this step, we apply TreeAttention to the input images at a lower resolution for computational efficiency. For training TreeAttention, we use binary cross entropy loss and ground truth derived from the bounding box annotations of the respective dataset. Since the goal of TreeAttention is to highlight the tree crown, we compute the alpha shape with $\alpha=100$ around the annotations of all apples in an image as ground truth. The alpha shape is a generalization of the convex hull and results in a tight-fitting hull around the bounding boxes. 
An example mask, which is used in the training of TreeAttention, is shown in~\cref{fig:alpha} with the corresponding bounding boxes depicted in~\cref{fig:boxes}. 


\subsection{Selective Tiling}
\label{sec:tiling}
To improve the detection of the small apples in high-resolution images, S$^3$AD follows a tiling approach~\cite{li2020density,ruuvzivcka2018fast,wilms2022localizing} and increases the relative size of the small apples for detection by extracting tiles. Subsequently, the object detector in S$^3$AD processes each tile in full resolution. Hence, our object detector can utilize all information available in the tiles and the small apples are less affected by downsampling in CNNs. Since standard tiling substantially increases the runtime, we utilize the attention maps generated by our TreeAttention module~(see~\cref{sec:tree_attention}) to select which tiles are processed by S$^3$AD's detector. 

Deriving the relevant tiles from an attention map, we first binarize the attention map with a threshold of $\tau=0.3$ and up-sample it to the original image size. Next, a $800\times800$ sliding window with a stride of 400 pixels is moved over the binarized attention map. For each window, the amount of pixels with an attention value greater than $\tau$ is determined, and the tile is selected for further processing in S$^3$AD if the amount is larger than $20\%$. The overlap between tiles ensures that each apple is fully contained in at least one tile. Compared to processing all tiles, this approach results in a significant reduction of processed tiles and inference time of S$^3$AD, without losing detection performance~(see~\cref{sec:evalAnalysisTiling}).

\subsection{Semi-supervised Detection}
\label{sec:detection}
S$^3$AD can be used together with an arbitrary object detector. Due to the success in previous crop detection applications~\cite{mu2020intact,bargoti2017deep,hani2020minneapple}, we choose Faster R-CNN~\cite{ren2015faster} with a Feature Pyramid Network (FPN) backbone~\cite{lin2017feature}. Faster R-CNN is a two-stage object detector that first generates class-agnostic object proposals, while later classifying and refining the proposals. In S$^3$AD, the classification is binary~(apple vs. background). The FPN backbone has been shown to improve detection results of Faster R-CNN on small objects~\cite{lin2017feature}, which fits the task of apple detection. As outlined above, we apply the detector to each selected tile, to improve the detection of small apples. Subsequently, the per-tile results are merged, as described in~\cref{sec:filtering}.


To utilize the large amount of unlabeled data in our proposed dataset MAD, we train S$^3$AD's Faster R-CNN detector in the semi-supervised Soft Teacher framework~\cite{xu2021end}. Soft Teacher utilizes two models for semi-supervised training: a teacher model and a student model. The teacher model learns to generate boxes from unlabeled data, while the student model is trained using a combination of labeled data and unlabeled data with boxes provided by the teacher model as pseudo ground truth. In this work, we use Faster R-CNN as our student model. 
Throughout the training, the teacher model is updated using the exponential moving average of the student model. Since the teacher generates thousands of box candidates, non-maximum suppression is applied to eliminate redundancy. Moreover, only candidates surpassing a given confidence score are considered as pseudo ground truth, aiming to reduce the number of false positives. We apply a confidence score threshold of 0.9 for the pseudo annotations. 

\subsection{Filtering \& Reconstruction}
\label{sec:filtering}
After applying the semi-supervised Faster R-CNN detector on each selected tile from an image, S$^3$AD reconstructs the result for the entire image by filtering and merging the per-tile results and applying an image-level filtering step. First, S$^3$AD filters the per-tile results to remove detections for partially visible apples along the tile borders that are fully visible in another tile. Such partial apple detections can cause false positives and removal after merging the per-tile results based on an IoU threshold is difficult, since detections of partial apples will naturally not have a high overlap with detections for the entire apple. To solve this problem, S$^3$AD removes the detections along a 100-pixel wide border around each tile prior to reconstructing the full image-level results. Note that we do not remove detections along the image edges.

After the per-tile filtering step, S$^3$AD merges the individual detection results and subsequently applies non-maximum suppression with an IoU threshold of 0.5.

\subsection{Training}
The semi-supervised Faster R-CNN detector in S$^3$AD is trained in the Soft Teacher framework and is initialized with the semi-supervised MS COCO weights. For training on MAD, we use tiles from all labeled training images and 41,415 unlabeled image tiles selected using TreeAttention~(see supp. mat. for details on selection and training of TreeAttention). Note that TreeAttention is trained as an individual system prior to training the actual detector.
During training, the image tiles are randomly resized in a range between 600 and 1600, and grouped in batches of size 10. We use a data sampling ratio of 0.2 in the Soft Teacher framework. Hence, four unlabeled images are randomly selected for each labeled tile in a batch. The model learns to minimize a joint loss consisting of a supervised and a weighted unsupervised loss term~(for more information, see~\cite{xu2021end}) and is trained for 100k steps using SGD with an initial learning rate, momentum, and weight decay of 0.001, 0.9, and 0.0001, respectively. Additionally, the learning rate is divided by 10 at 60k and 80k steps and the data sampling ratio is gradually decreased to 0 in the last 5,500 steps. To ensure high-quality pseudo annotations, the foreground threshold for using the teacher's pseudo ground truth is set to 0.9.

\section{Evaluation}
\label{sec:eval}
To validate the strength of S$^3$AD, we conduct an evaluation on our introduced dataset MAD. As outlined in~\cref{sec:data_acqui}, MAD covers complex orchard environments with apples of small relative size and only a limited amount of annotated training images. Additionally, we assess the generality of our approach w.r.t. other apple detection datasets by training and testing on the MSU dataset~\cite{CHU2023100284}. All results in this chapter are generated on the respective test splits.

On MAD, we compare our S$^3$AD to five strong fully-supervised baselines that specifically address the detection of small objects: Faster R-CNN~\cite{ren2015faster} with FPN backbone~\cite{lin2017feature}, which is frequently used in crop detection approaches~\cite{mu2020intact,bargoti2017deep,hani2020minneapple}, PANet~\cite{liu2018path}, SNIPER~\cite{singh2018sniper}, AutoFocus~\cite{najibi2019autofocus}, and Deformable DETR~\cite{zhudeformable}, a transformer-based object detector. A direct comparison to other crop detection approaches is impossible, due to the lack of publicly available implementations. For training the fully-supervised baselines, we only utilize the 66 annotated training images of MAD, containing 10,089 annotated apples.  On MSU, we compare to the base system of S$^3$AD, which is Faster R-CNN+FPN. A comparison to O2RNet~\cite{CHU2023100284} proposed by the authors of MSU is not possible due to the lack of an available implementation. To match the semi-supervised setup of MAD, we train S$^3$AD and Faster R-CNN+FPN with annotations on only $10\%$ of the training image, resulting in 93 annotated training images. Note that in MAD, only $1.5\%$ of the training images are labeled.

To assess the quality of the apple detection results, we utilize the commonly used measures Average Recall~(AR) and Average Precision~(AP) based on 100 detections. While AR evaluates how many apples are found and how precisely they are located, AP also considers the false positives. Both metrics are regularly used for evaluating object detection~\cite{ren2015faster,redmon2018yolov3} and crop detection approaches~\cite{koirala2019deep,hani2020minneapple,stein2016image,wilms2022localizing}. Supplementing these quantitative results, we also provide qualitative results.

\subsection{Results on MAD}
\label{sec:MADResults}

\subsubsection{Overall Quantitative Results}
\label{sec:MADquanRes}

\begin{table*}
\centering
\caption{Quantitative results in terms of Average Precision (AP) and Average Recall (AR) for S$^3$AD with and without our selective tiling, as well as five object detectors. The results are generated on the test split of our proposed dataset MAD.}
\label{tab:apples_ap_ar}
\begin{tabular}{lccccc}
\toprule
System &  Backbone & Semi-supervised & Selective Tiling & AP$\uparrow$ & AR$\uparrow$\\ \hline  
SNIPER~\cite{singh2018sniper}&ResNet-101 & & & 0.387 &  0.419 \\
AutoFocus~\cite{najibi2019autofocus}&ResNet-101 & & & 0.417 &  0.419 \\
\midrule
Faster R-CNN+FPN~\cite{ren2015faster,lin2017feature} &ResNet-50 & & &  0.368 & 0.408 \\ 
PANet~\cite{liu2018path}&ResNet-50 & & & 0.378 &  0.418 \\ 
Deformable DETR~\cite{zhudeformable}&ResNet-50 & & & 0.385 &  0.438 \\ 
\midrule
S$^3$AD w/o tiling (ours) &ResNet-50 & \cmark  & &  0.408 & 0.448  \\ 
S$^3$AD (ours) &ResNet-50 & \cmark & \cmark & \textbf{0.423} & \textbf{0.467}  \\ \bottomrule
\end{tabular}
\end{table*}

Table~\ref{tab:apples_ap_ar} presents the quantitative results of S$^3$AD with and without selective tiling and the five fully-supervised baselines. From the results, it is clearly visible that we outperform all baselines. Compared to Faster R-CNN+FPN, the base system of S$^3$AD, the improvement is $10.9\%$ in terms of AP by utilizing unlabeled data and even $14.9\%$ by  additionally introducing selective tiling. Compared to the other dedicated small object detection systems PANet and SNIPER, the improvement is up to $11.9\%$ in terms of AP. S$^3$AD even outperforms the transformer-based system Deformable DETR by $9.9\%$. Only AutoFocus is able to stay competitive with S$^3$AD. Here the improvement is only $1.4\%$ in terms of AP, but $11.5\%$ in terms of AR, highlighting the strong recall of S$^3$AD. Moreover, AutoFocus is trained with a ResNet-101 backbone, while  S$^3$AD is trained with a smaller ResNet-50 backbone. Finally, comparing S$^3$AD with and without tiling indicates the positive influence of our selective tiling on the results, with an improvement of $3.7\%$. Overall, our results show the strong positive effect of utilizing unlabeled data in the semi-supervised training and the benefit of our selective tiling.

\begin{figure*}[tb]
        \centering
        \begin{subfigure}[b]{0.31\textwidth}   
            \centering 
            \includegraphics[width=\linewidth]{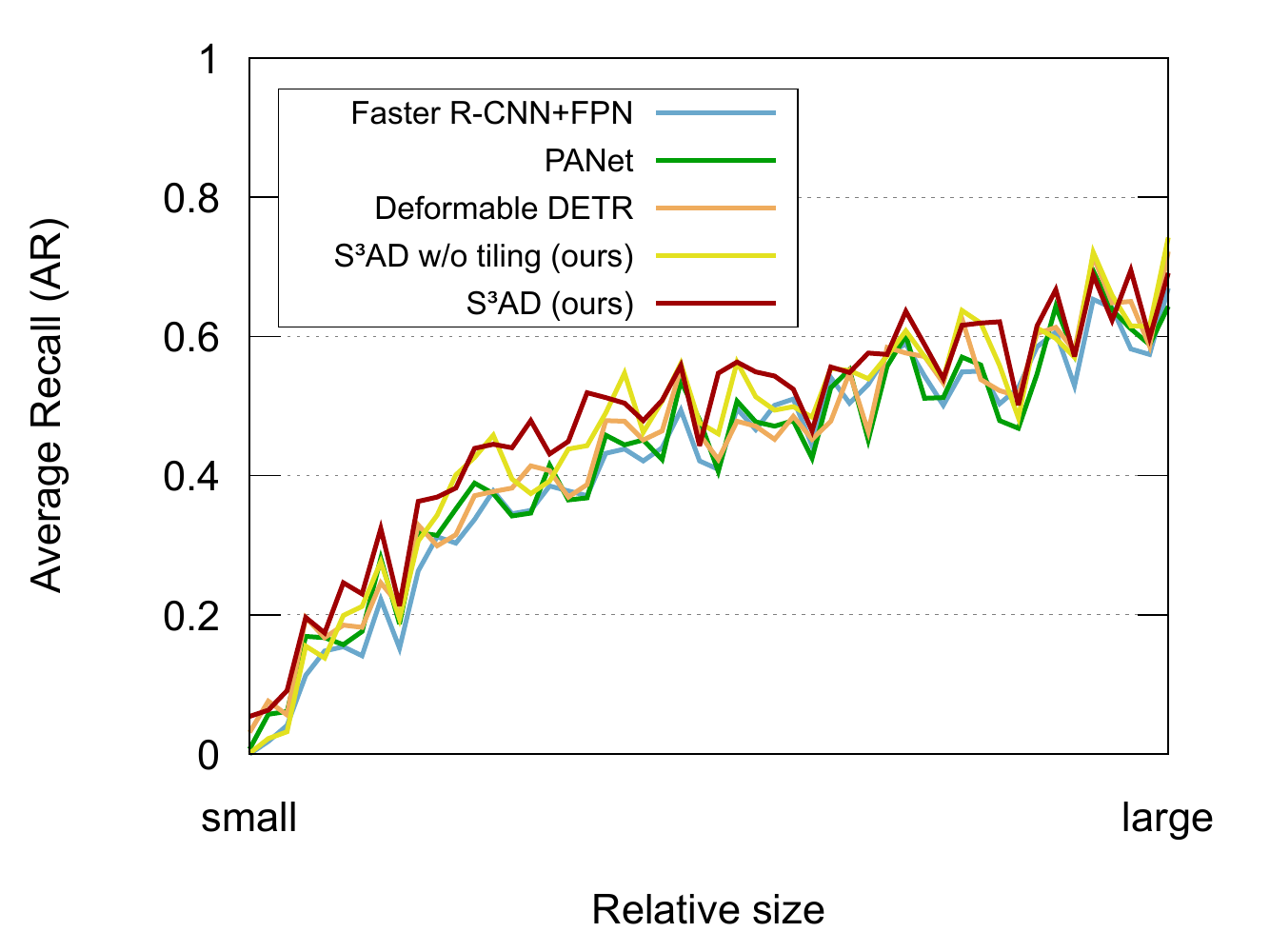}
            \caption{Relative size}
            \label{fig:propEval_relSize}
        \end{subfigure}
        ~
        \begin{subfigure}[b]{0.31\textwidth}   
            \centering 
            \includegraphics[width=\linewidth]{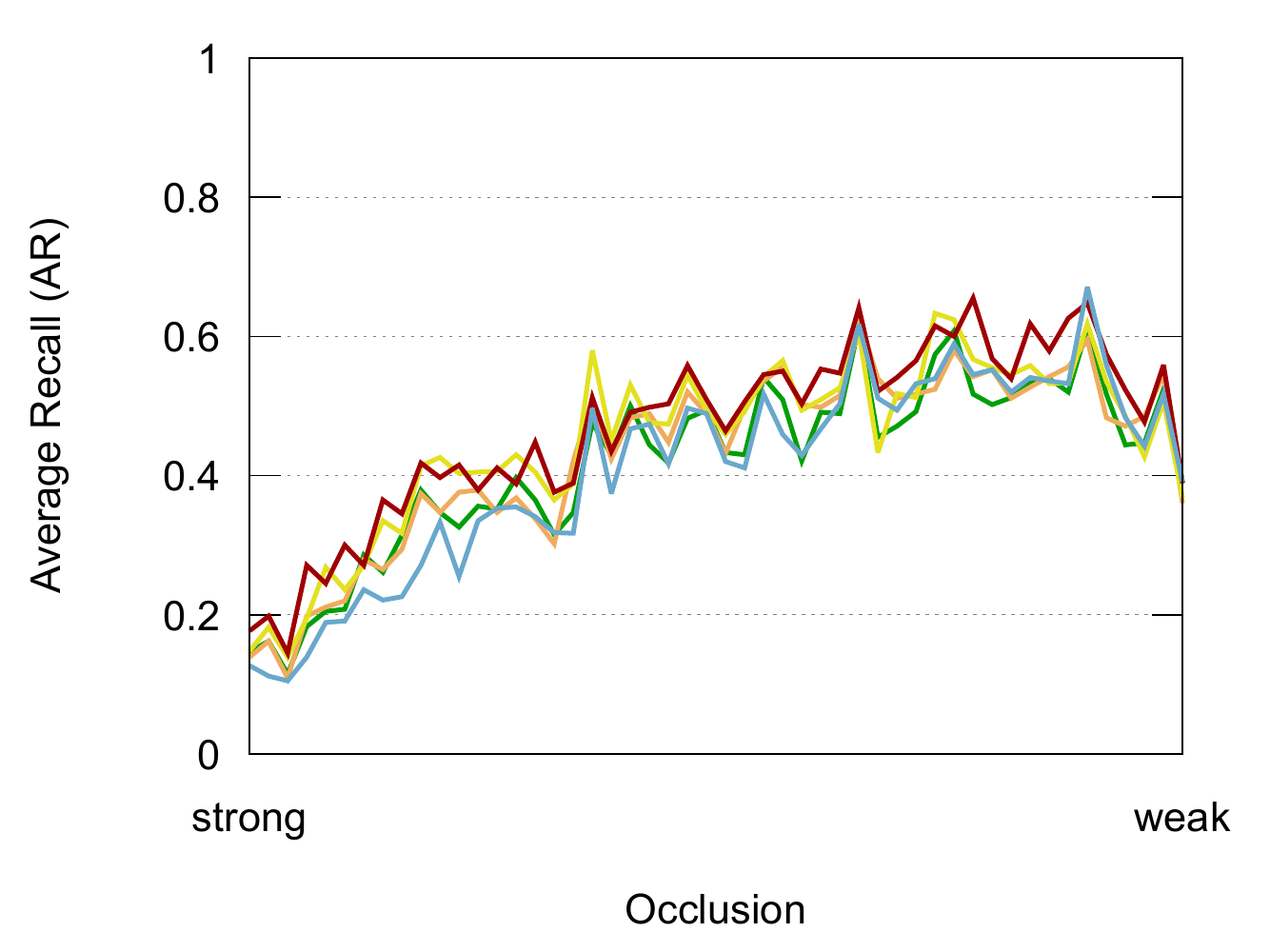}
            \caption{Level of occlusion}
            \label{fig:propEval_occ}
        \end{subfigure}
        ~
        \begin{subfigure}[b]{0.31\textwidth}   
            \centering 
            \includegraphics[width=\linewidth]{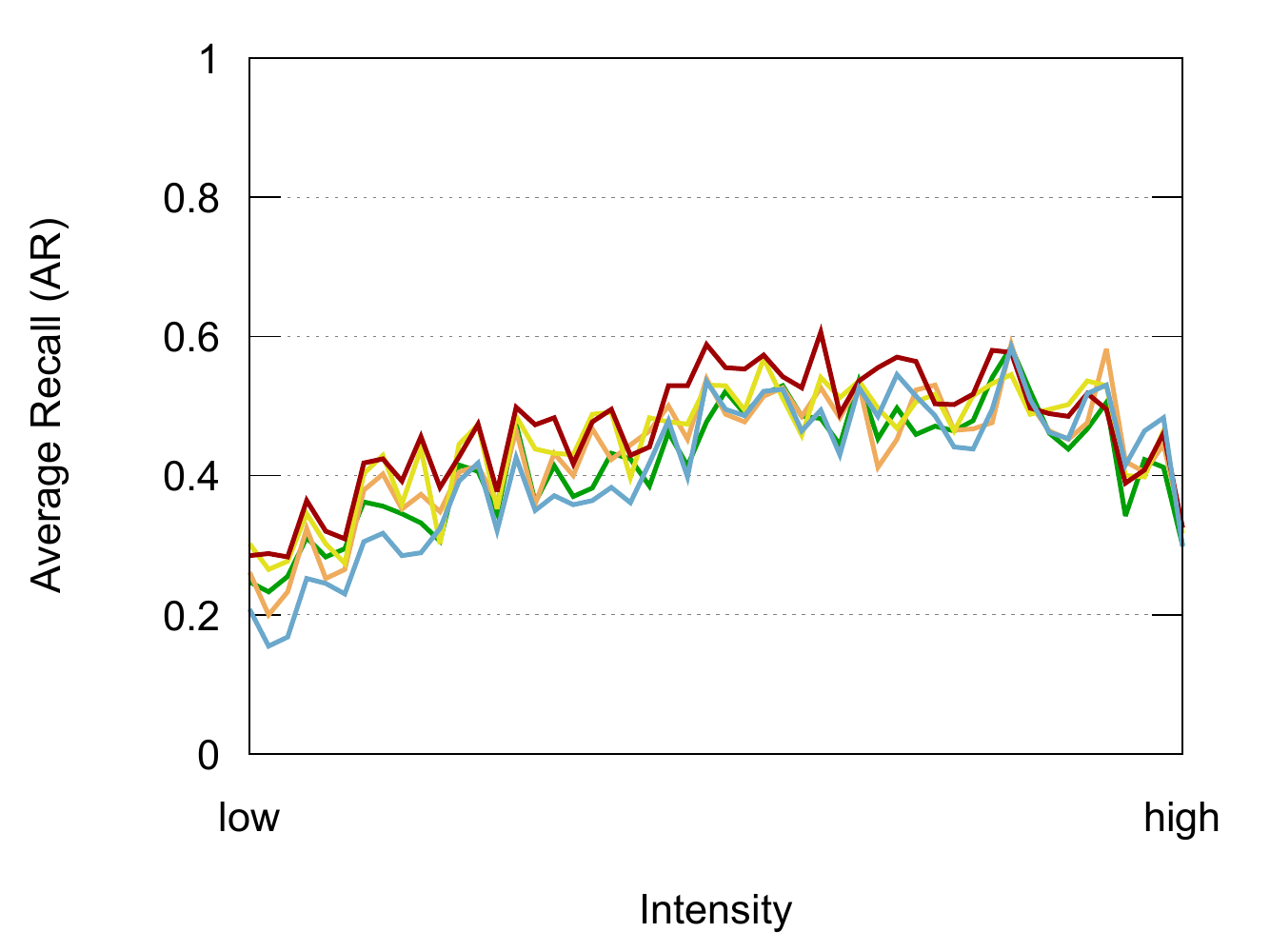}
            \caption{Lighting condition}
            \label{fig:propEval_inten}
        \end{subfigure}
        
        \caption{Apple detection results of S$^3$AD with and without tiling, Faster R-CNN+FPN, PANet, and Deformable DETR in terms of AR for property-specific ranges. Each point on the curves represents a bin of $2\%$ of the annotated apples in the test split of our dataset MAD.}
        \label{fig:propEval}
    \end{figure*}
    
\subsubsection{Property-based Quantitative Results}
\label{sec:evalProp}
After discussing the overall quantitative results, we will examine the behavior of the previously discussed detection systems based on a ResNet-50 backbone: S$^3$AD with and without our selective tiling, S$^3$AD's base system Faster R-CNN+FPN, PANet, and Deformable DETR. We examine the behavior w.r.t. three apple properties annotated in MAD: relative size, level of occlusion, and lighting conditions. The AR-based results for different levels of each property are given in~\cref{fig:propEval}, where each data point represents $2\%$ of the annotated apples on property-specific scales. We chose AR here, to show how many apples featuring the respective property are found without penalizing duplicates. 

Analyzing the results w.r.t. the relative size of objects, it is clearly visible that S$^3$AD almost consistently outperforms all other systems. Especially on small objects~(left part in~\cref{fig:propEval_relSize}), there is a stronger improvement over the other systems. Analyzing this improvement in more detail, we quantify the average improvement across the first third of the data points/bins~(small objects) and the last third~(large objects). S$^3$AD with our tiling outperforms S$^3$AD without tiling by $15\%$ on small objects, while the improvement on large objects is only $1.7\%$. Similarly, the improvement over Faster R-CNN+FPN on small objects is $35.3\%$, while being only $7.4\%$ on large objects. Despite the improvements, all system's results degrade with decreasing relative object size.

On the other two annotated properties, level of occlusion and lighting conditions, the systems behave similarly to each other. While S$^3$AD almost consistently outperforms the other systems, all systems struggle with dark or bright apples as well as strongly occluded apples similarly. While the results of S$^3$AD degrade from weakly occluded apples (right third in~\cref{fig:propEval_occ}) to strongly occluded apples~(left third in~\cref{fig:propEval_occ}) by $42.3\%$, the difference is $45.4\%$ for PANet and $46.7\%$ for Deformable DETR. Similarly, moving from well-illuminated apples~(central third in~\cref{fig:propEval_inten}) to dark apples~(left third in~\cref{fig:propEval_inten}), the drop in performance is $24.4\%$ for S$^3$AD and $26.5\%$ for PANet.

Overall, this detailed evaluation highlights the positive influence of our proposed selective tiling in S$^3$AD on the detection of small apples and the generally improved performance of S$^3$AD on small apples compared to other systems. Moreover, from all three properties, we can derive and quantify challenging conditions for apple detection systems: small apples, strongly occluded apples, and dark or bright apples.

\subsubsection{Qualitative Results}
\label{sec:qualRes}

\begin{figure*}[t]
\centering
\begin{tabular}{cccc}
\includegraphics[width = .22\linewidth]{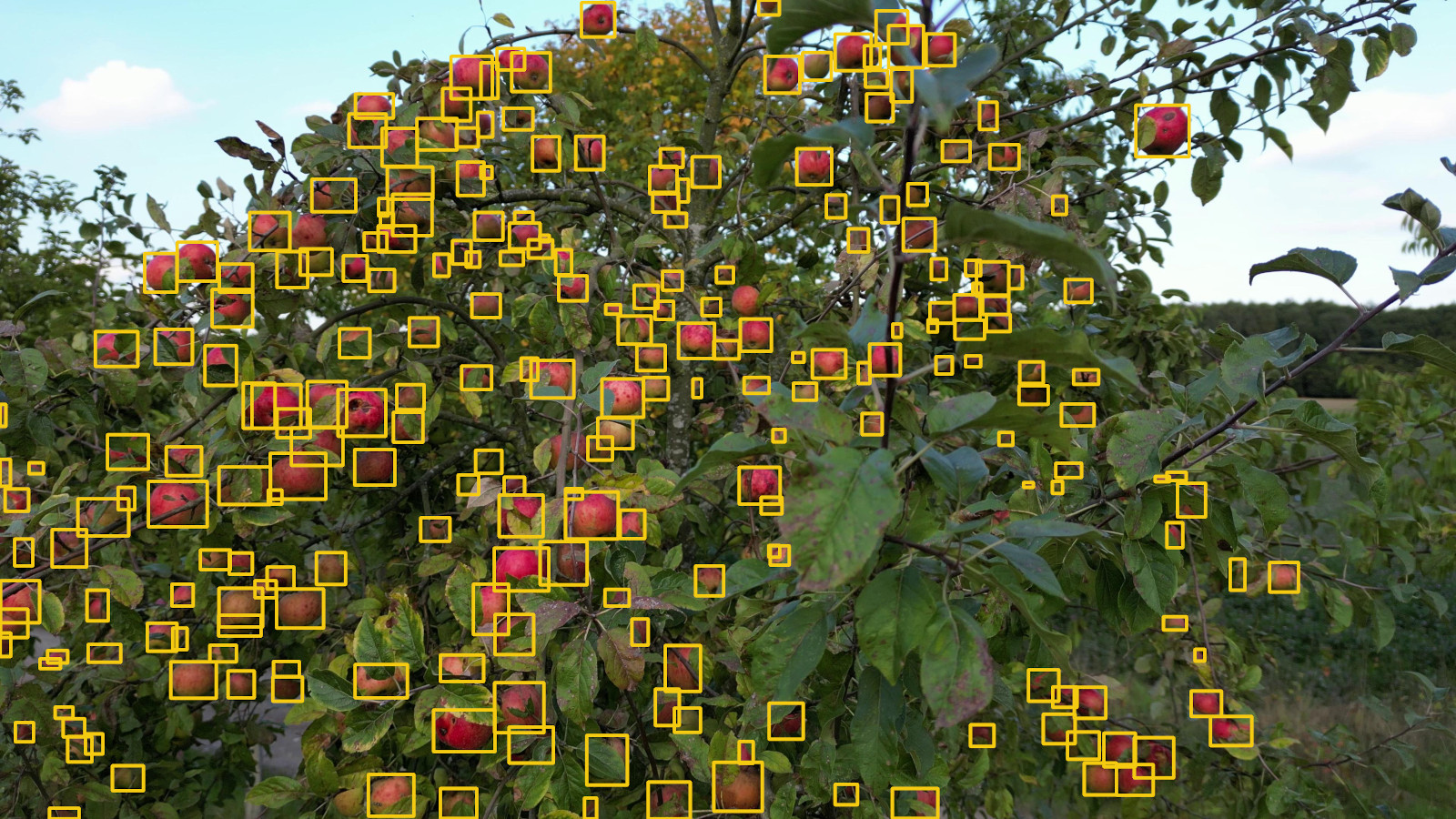} &
\includegraphics[width = .22\linewidth]{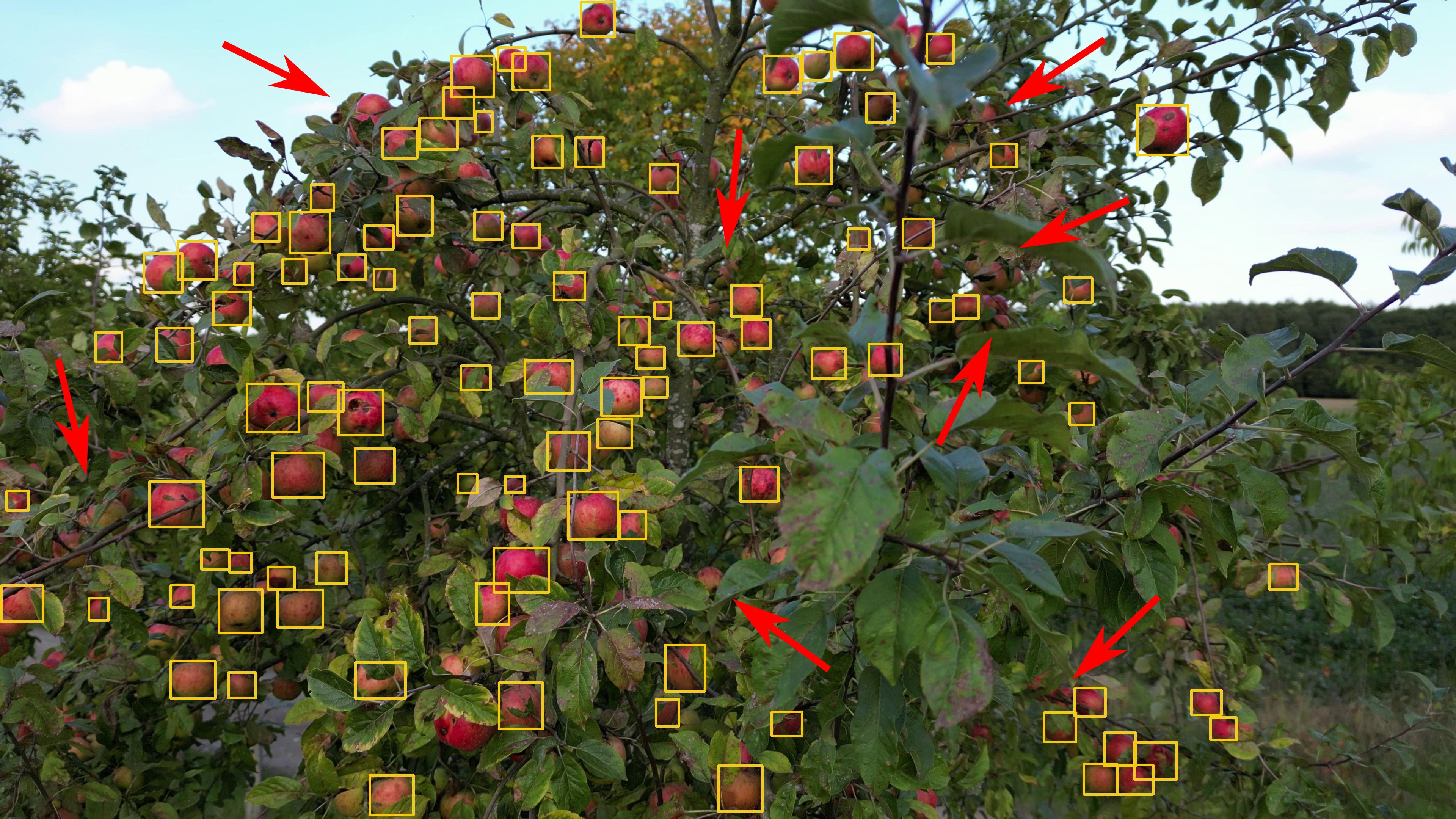}&
\includegraphics[width = .22\linewidth]{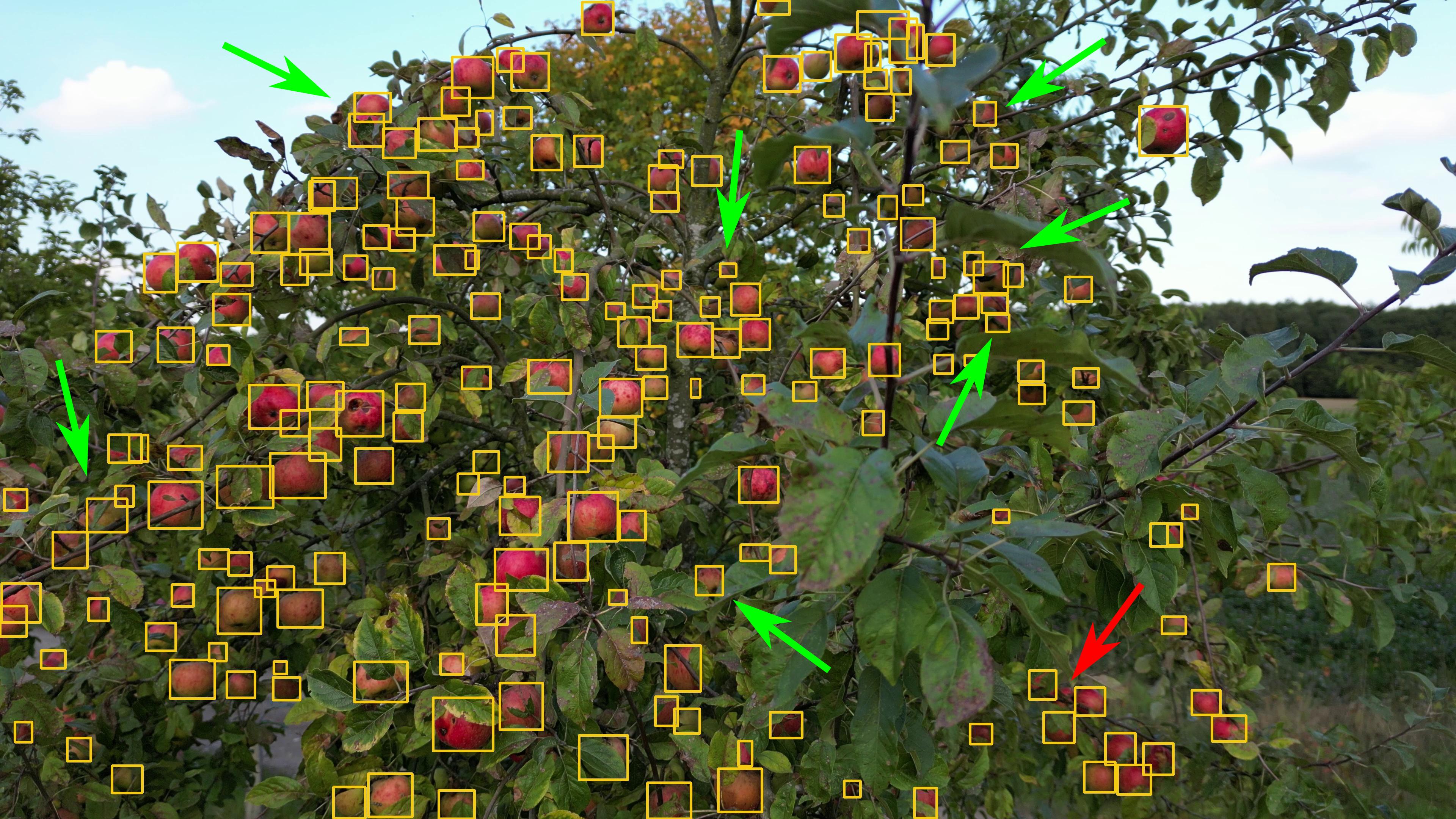} &
\includegraphics[width = .22\linewidth]{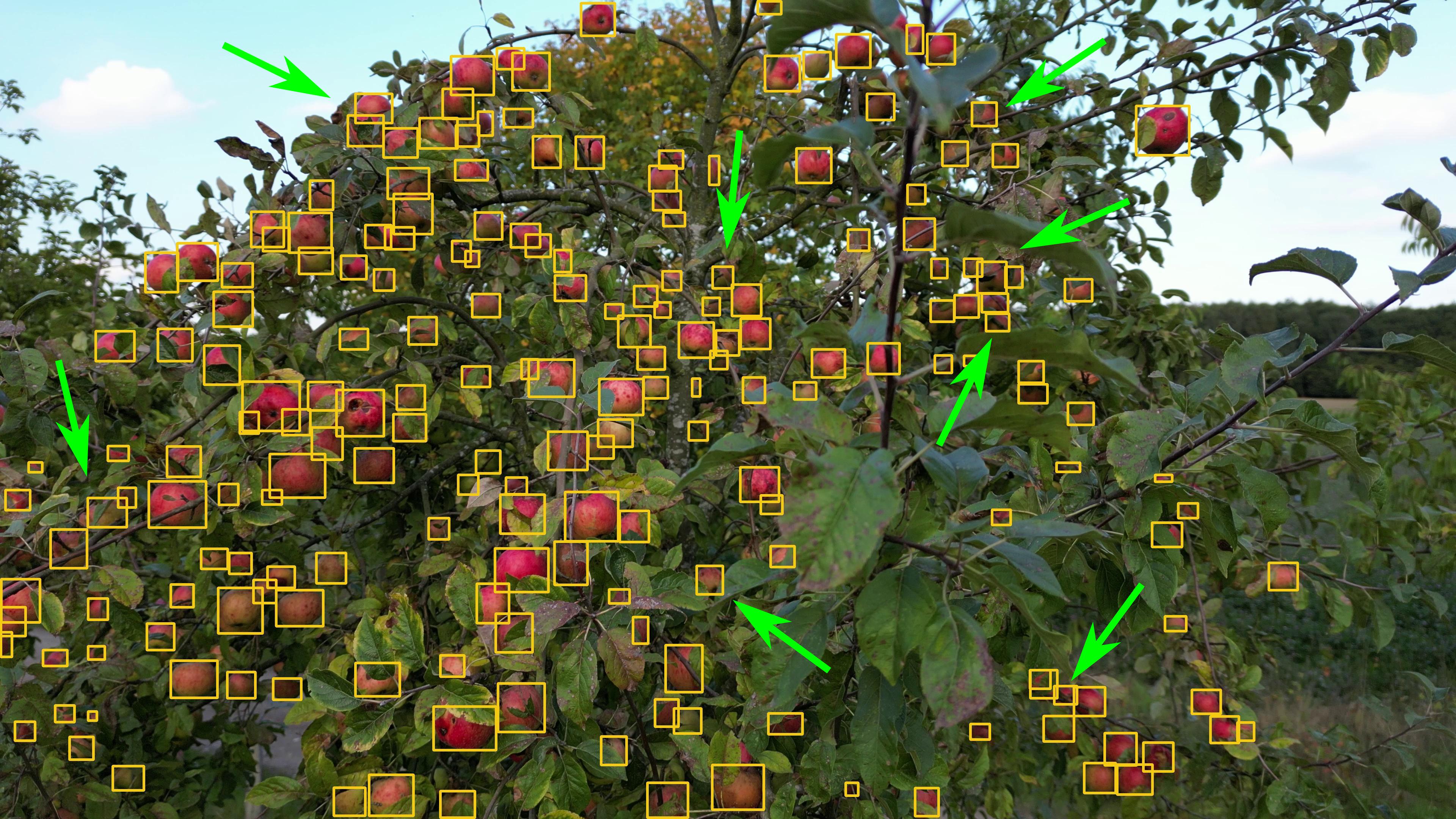}\\
\includegraphics[width = .22\linewidth]{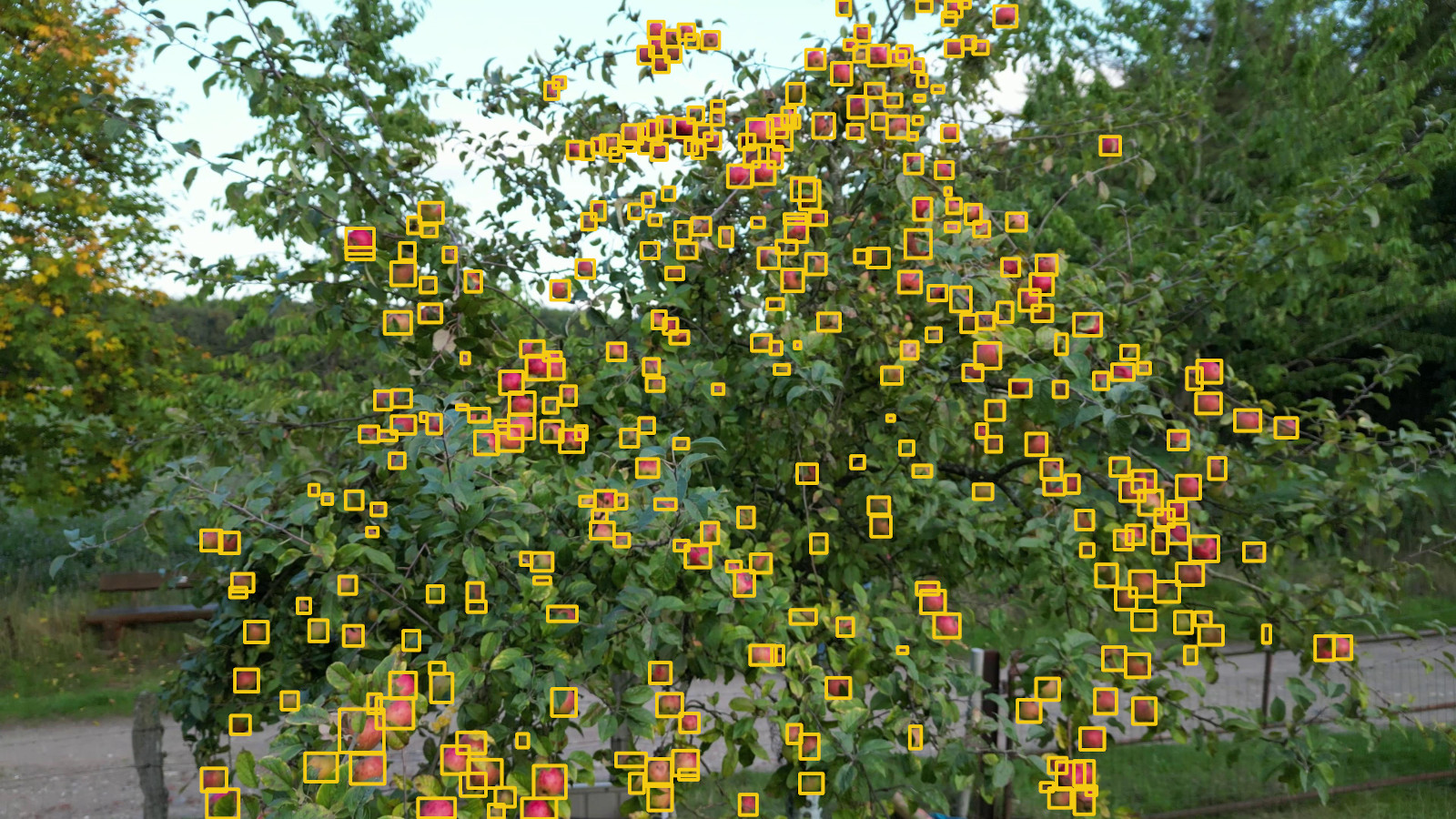} &
\includegraphics[width = .22\linewidth]{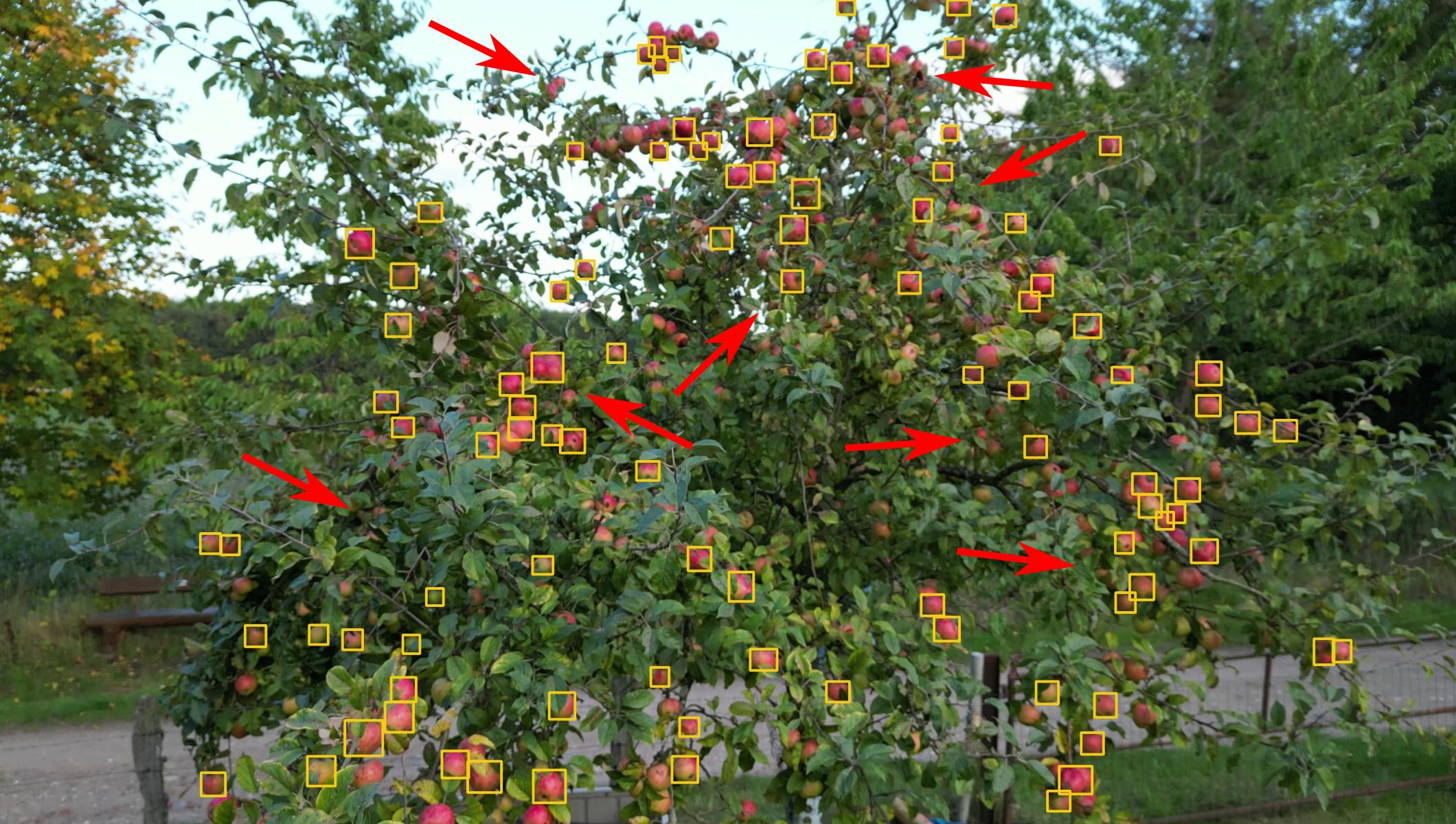}&
\includegraphics[width = .22\linewidth]{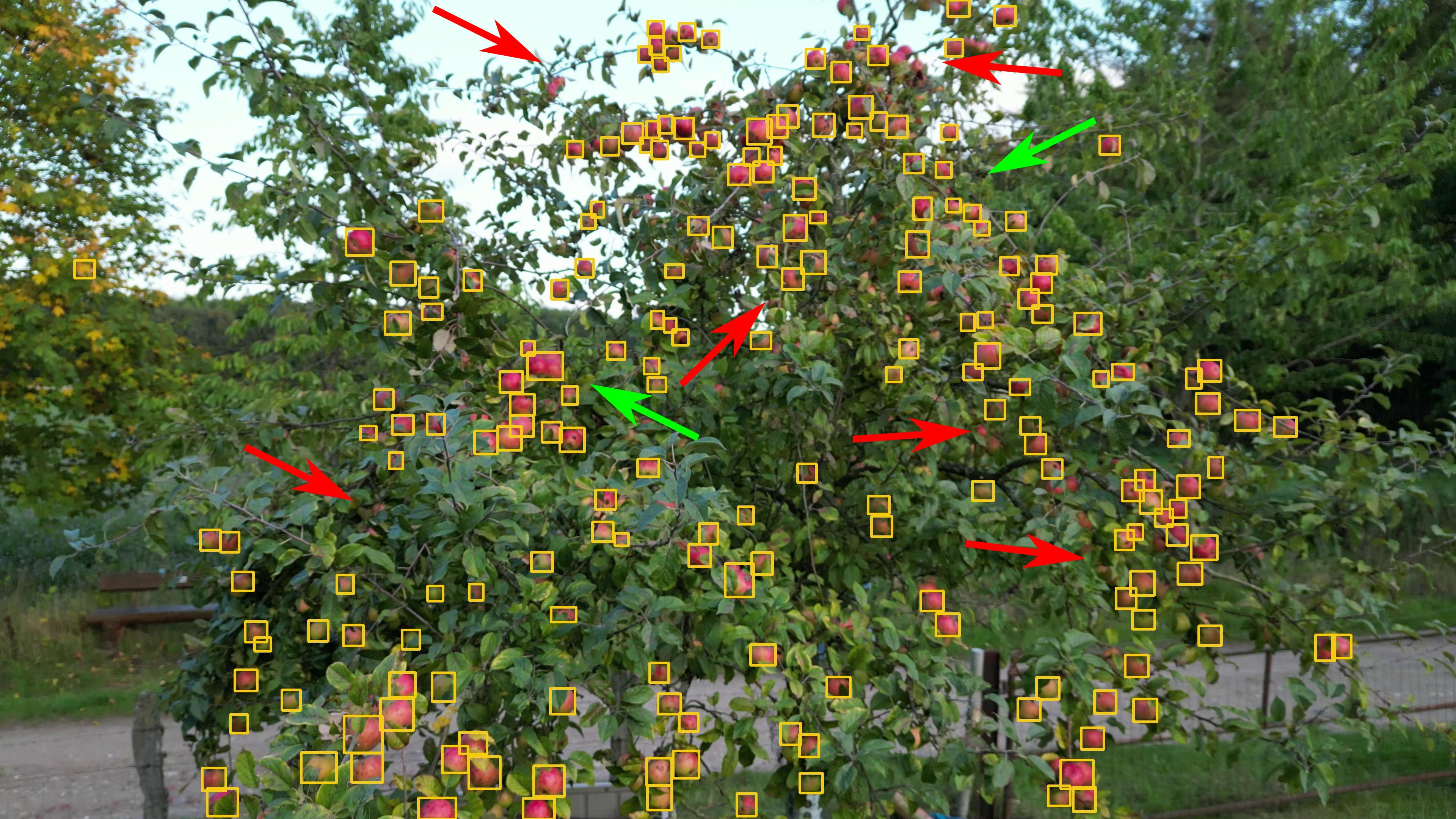} &
\includegraphics[width = .22\linewidth]{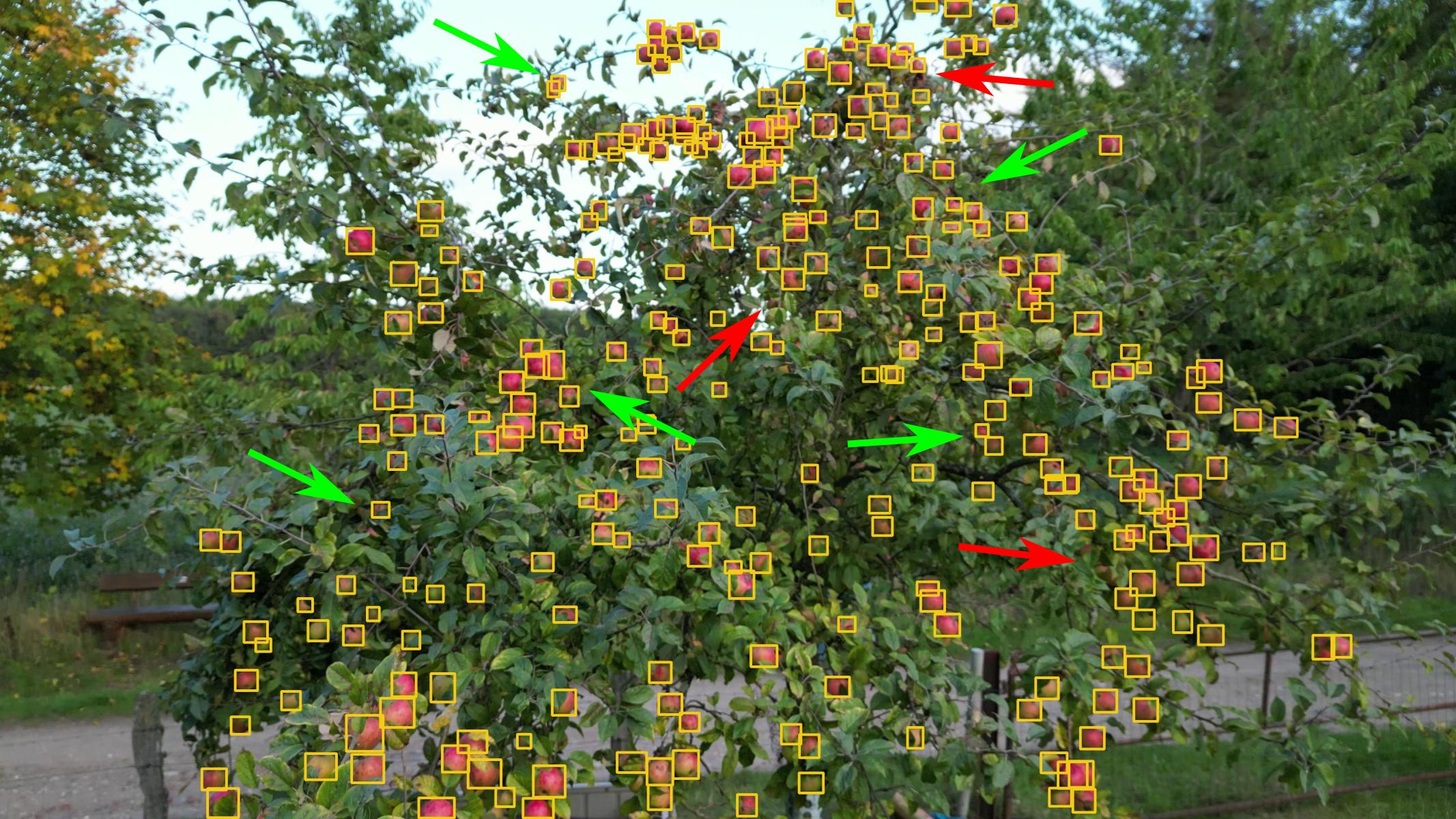}\\
{\footnotesize Input image with ground truth}&{\footnotesize Faster R-CNN+FPN}&{\footnotesize AutoFocus}&{\footnotesize S$^3$AD (ours)}
\end{tabular}
\caption{Qualitative results of S$^3$AD, Faster R-CNN+FPN, and AutoFocus on the test split of our dataset MAD. Red arrows denote missed apples, while green arrows denote the detection of such apples by other systems.
}
\label{fig:qual_semi}
\end{figure*}

\begin{figure}
\centering
\begin{tabular}{cc}
\includegraphics[width = .45\linewidth]{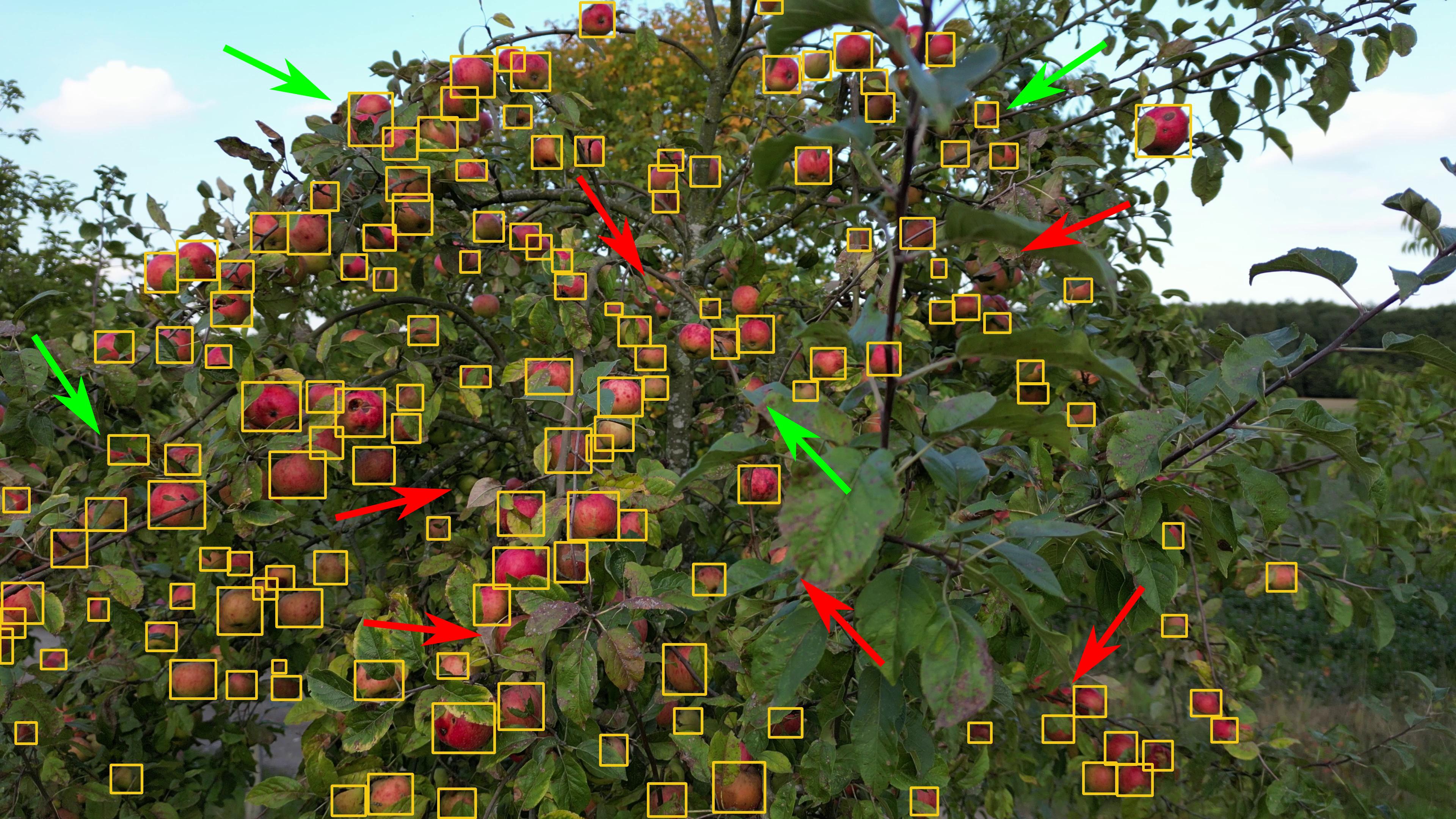} &
\includegraphics[width = .45\linewidth]{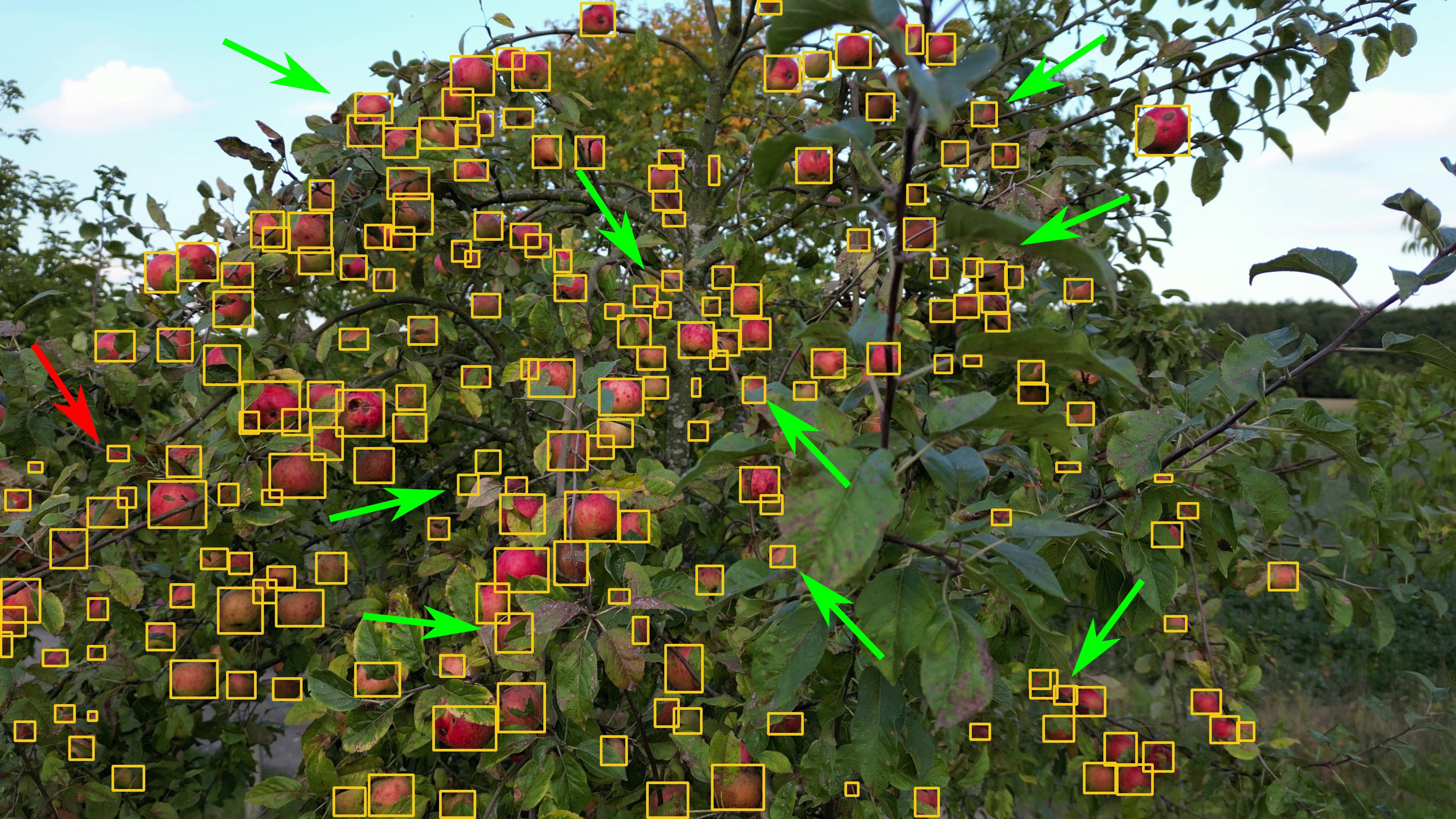}\\
\includegraphics[width = .45\linewidth]{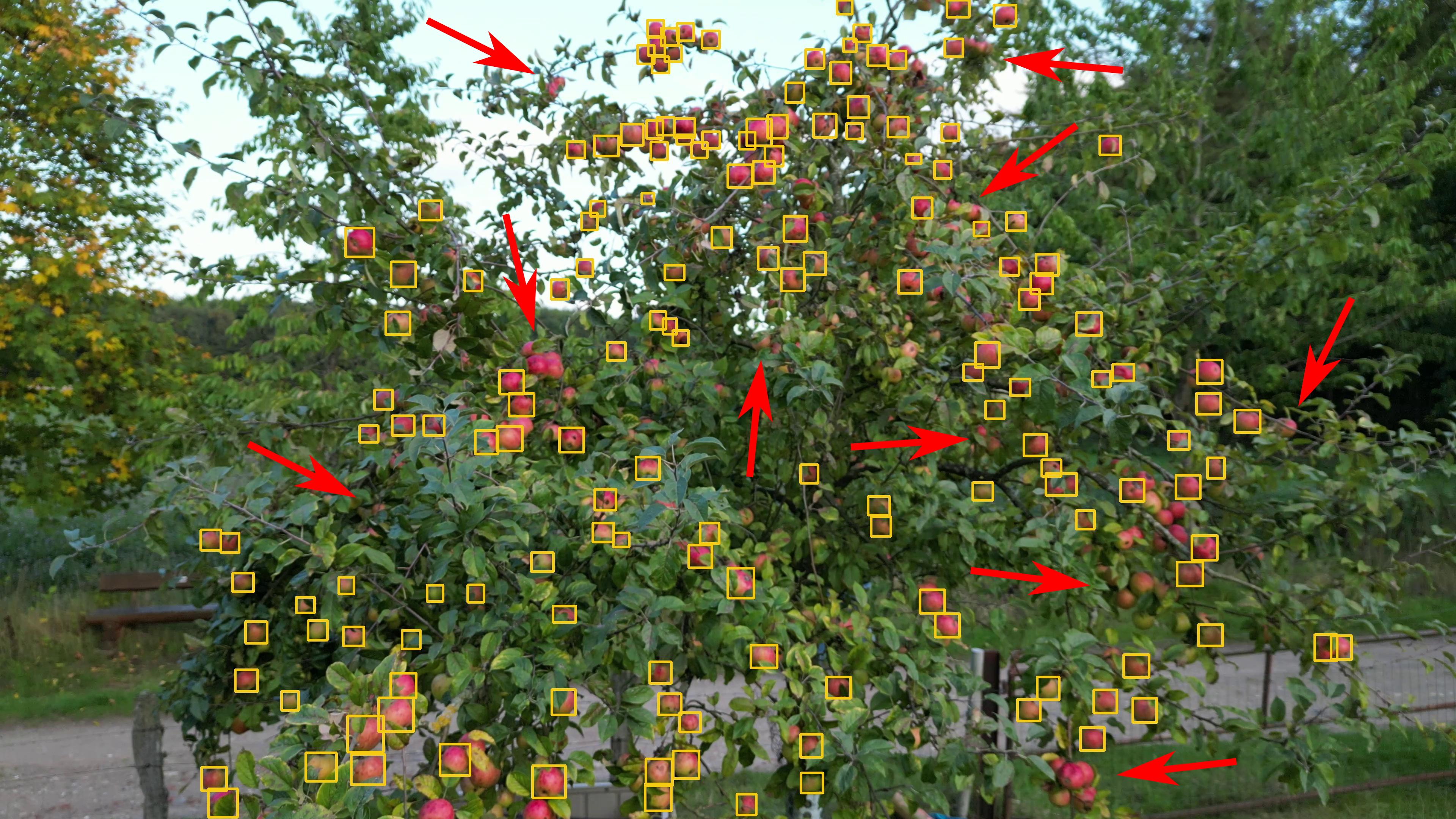} &
\includegraphics[width = .45\linewidth]{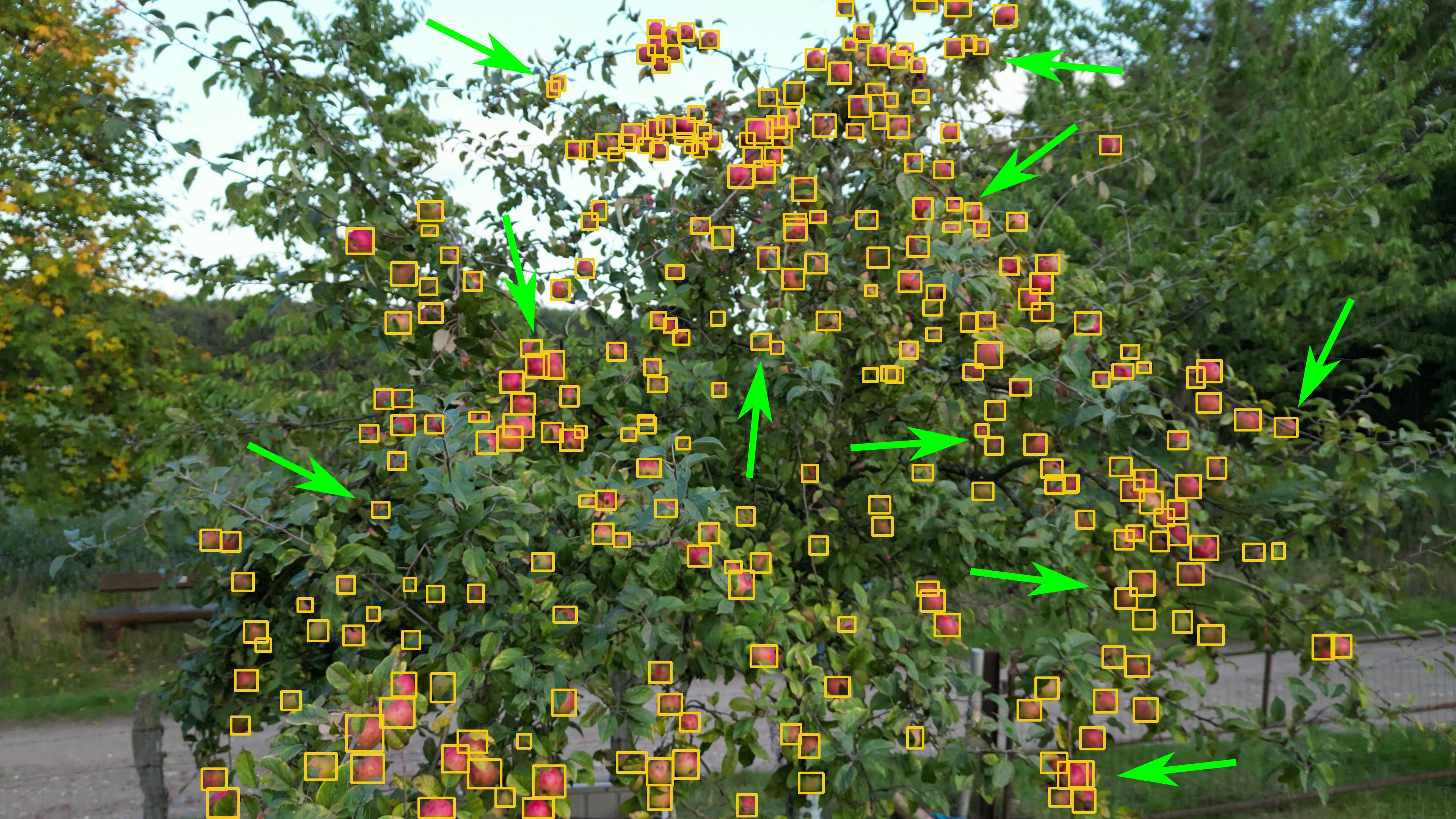}\\
{\footnotesize No tiling}&{\footnotesize Selective tiling (ours)}
\end{tabular}
\caption{Qualitative results of S$^3$AD with and without using selective tiling to improve the detection of small apples on the test split of our MAD dataset. See~\cref{fig:qual_semi} for the input images with ground truth. Red arrows denote missed apples, while green arrows denote the detection of such apples by other systems.
}
\label{fig:qual_tiling}
\end{figure}

Besides the quantitative results, we also present qualitative results in~\cref{fig:qual_semi} and~\cref{fig:qual_tiling}. The results in~\cref{fig:qual_semi} indicate that S$^3$AD is able to improve the detection of small apples compared to Faster R-CNN+FPN and AutoFocus. For instance, in the upper example in ~\cref{fig:qual_semi} S$^3$AD detects the small apple in the top left corner that Faster R-CNN+FPN misses. Moreover, several very small apples that are partially occluded are better detected by S$^3$AD compared to Faster R-CNN+FPN and AutoFocus~(see arrows). This also confirms the findings from~\cref{sec:evalProp} about the strong performance of S$^3$AD on small apples and the generally better results of S$^3$AD across all levels of occlusions. The lower example in~\cref{fig:qual_semi} also supports these findings. However, in this example, it is also clearly visible that even S$^3$AD still misses a few small apples, mainly due to occlusion~(central red arrow) or insufficient lighting conditions~(bottom right arrow). All three properties were also identified as major challenges for apple detection in~\cref{sec:evalProp}.

Focusing on the influence of the selective tiling, \cref{fig:qual_tiling} shows the results of S$^3$AD with and without the selective tiling. As expected from the quantitative results, several small apples are missed without the selective tiling denoted by the red arrows. Multiple such examples are visible in the upper part and the right part of the lower example. For instance, the four small apples in the bottom right of the lower example, marked by an arrow, are all missed without our selective tiling. Across both examples, several small apples are missed due to strong occlusion. This is also slightly improved by applying the selective tiling, as the results in the top right of the upper example and the quantitative results in~\cref{fig:propEval_occ} indicate. Still, several strongly occluded apples are still missed despite selective tiling.

Overall, the qualitative results show the strong performance of our system and the substantial improvement using our selective tiling approach for the detection of small apples and to some extent also to strongly occluded apples. 



\subsection{Results on MSU}
\label{sec:MSUResults}
The results in the MSU dataset for our S$^3$AD and its base system Faster R-CNN+FPN are presented in~\cref{tab:resultsMSU}. Generally, the tendency of the results is similar to MAD, with S$^3$AD outperforming Faster R-CNN+FPN. However, the margin between the methods is much larger on MSU compared to MAD. While the improvement in terms of AP from Faster R-CNN+FPN to S$^3$AD was $14.9\%$ on MAD, it is now up to $50.5\%$. To examine this effect in more detail, we also compare to S$^3$AD without tiling to disentangle the effect of our selective tiling and the semi-supervised training. S$^3$AD without tiling already outperforms Faster R-CNN+FPN by $44.2\%$ in terms of AP, indicating the strong influence of semi-supervised learning in the absence of a large amount of annotated images. Still, the proposed selective tiling is effective on the MSU dataset as well, leading to another $4.4\%$ improvement when added to S$^3$AD, which is similar to the results on MAD. Comparing S$^3$AD with selective tiling to a version of Faster R-CNN+FPN trained on the entire annotated training dataset of MSU (842 annotated images), reveals that despite using only around $11\%$ of the annotated data, the result of S$^3$AD reaches $90\%$ of the fully supervised Faster R-CNN results (0.518 vs. 0.579). This highlights again the strength of the proposed semi-supervised training in S$^3$AD. Overall, the results on MSU show that S$^3$AD generalizes well to other apple detection datasets, while again displaying the strength of the semi-supervised training. 
\begin{table}
\centering
\caption{Quantitative results in terms of Average Precision (AP) and Average Recall (AR) for S$^3$AD with and without tiling, as well as Faster R-CNN+FPN on the test split of MSU.}
\label{tab:resultsMSU}
\begin{tabular}{lcc}
\toprule
System &   AP$\uparrow$ & AR$\uparrow$\\ \hline  
Faster R-CNN+FPN~\cite{ren2015faster,lin2017feature} &  0.344 & 0.439 \\ 
\midrule
S$^3$AD w/o tiling (ours) & 0.496 & 0.574  \\ 
S$^3$AD (ours) & \textbf{0.518} & \textbf{0.594}  \\ \bottomrule
\end{tabular}
\end{table}

\subsection{Analysis of Selective Tiling Approach}
\label{sec:evalAnalysisTiling}
After presenting the main results of our system, we further analyze our selective tiling approach, which improves the detection of small apples, as shown in~\cref{fig:propEval_relSize}.

\subsubsection{Quality of TreeAttention}
\begin{figure}
\centering
\begin{tabular}{cc}
\includegraphics[width = .45\linewidth]{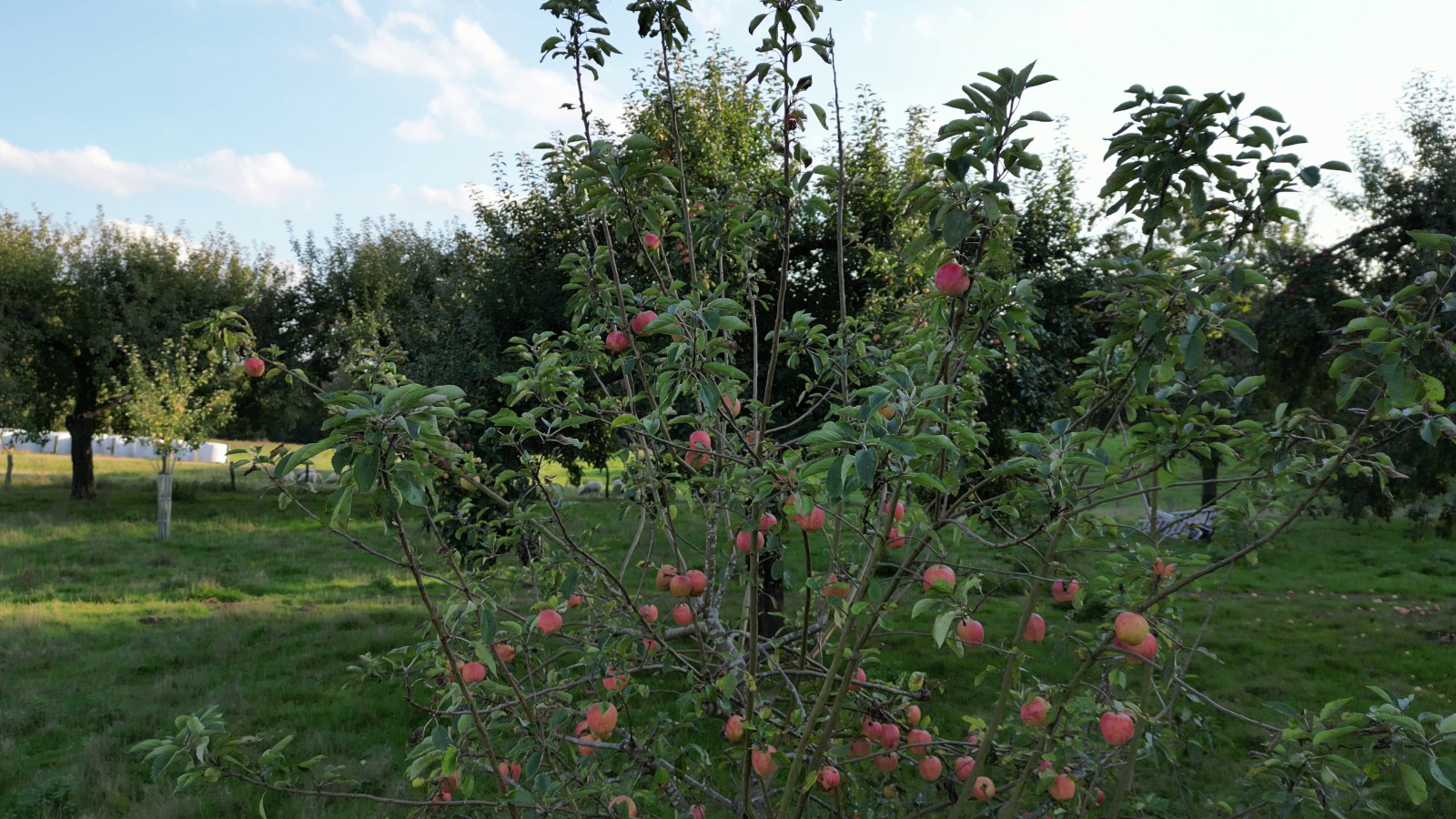} &
\includegraphics[width = .45\linewidth]{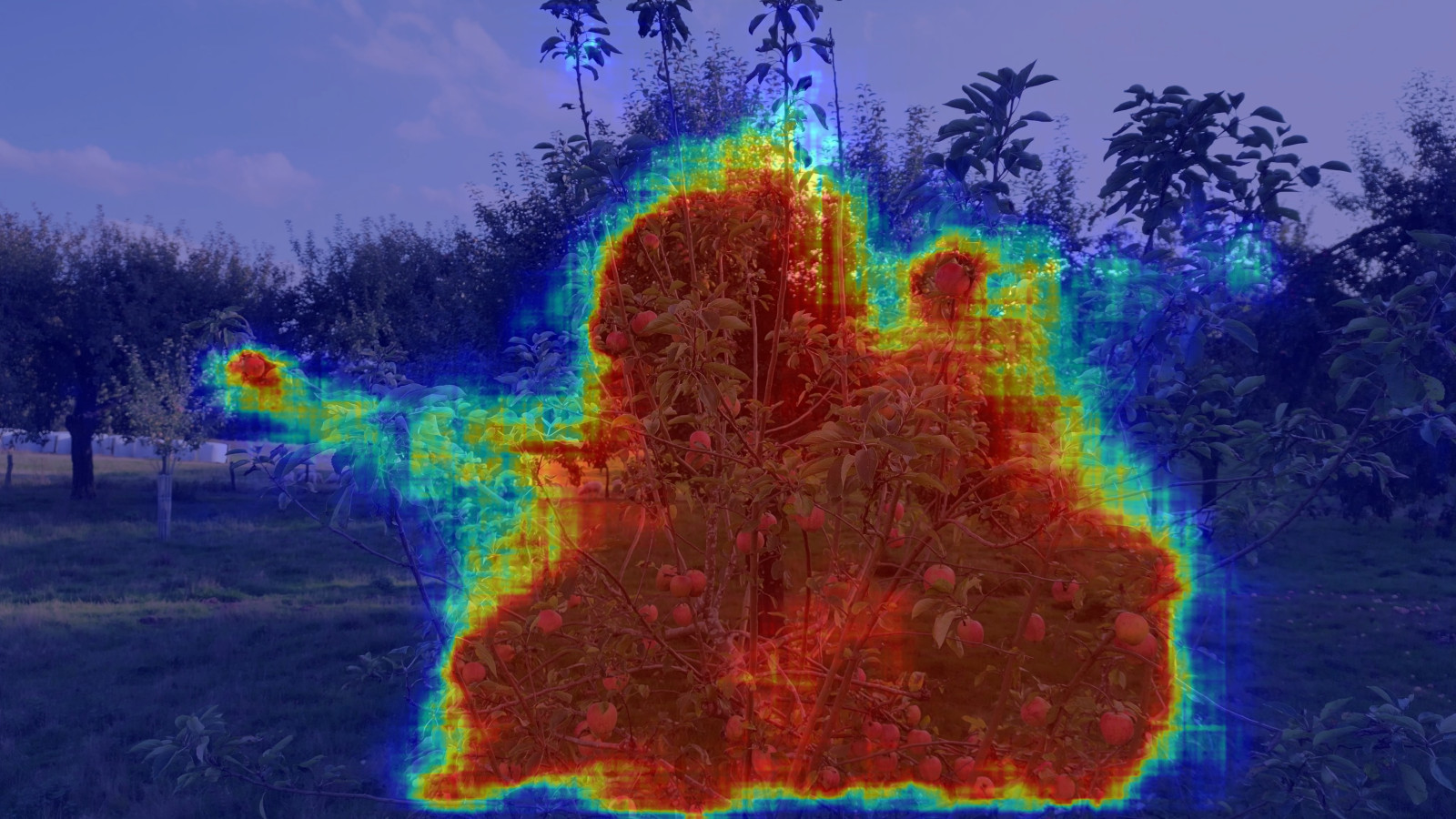}\\
\includegraphics[width = .45\linewidth]{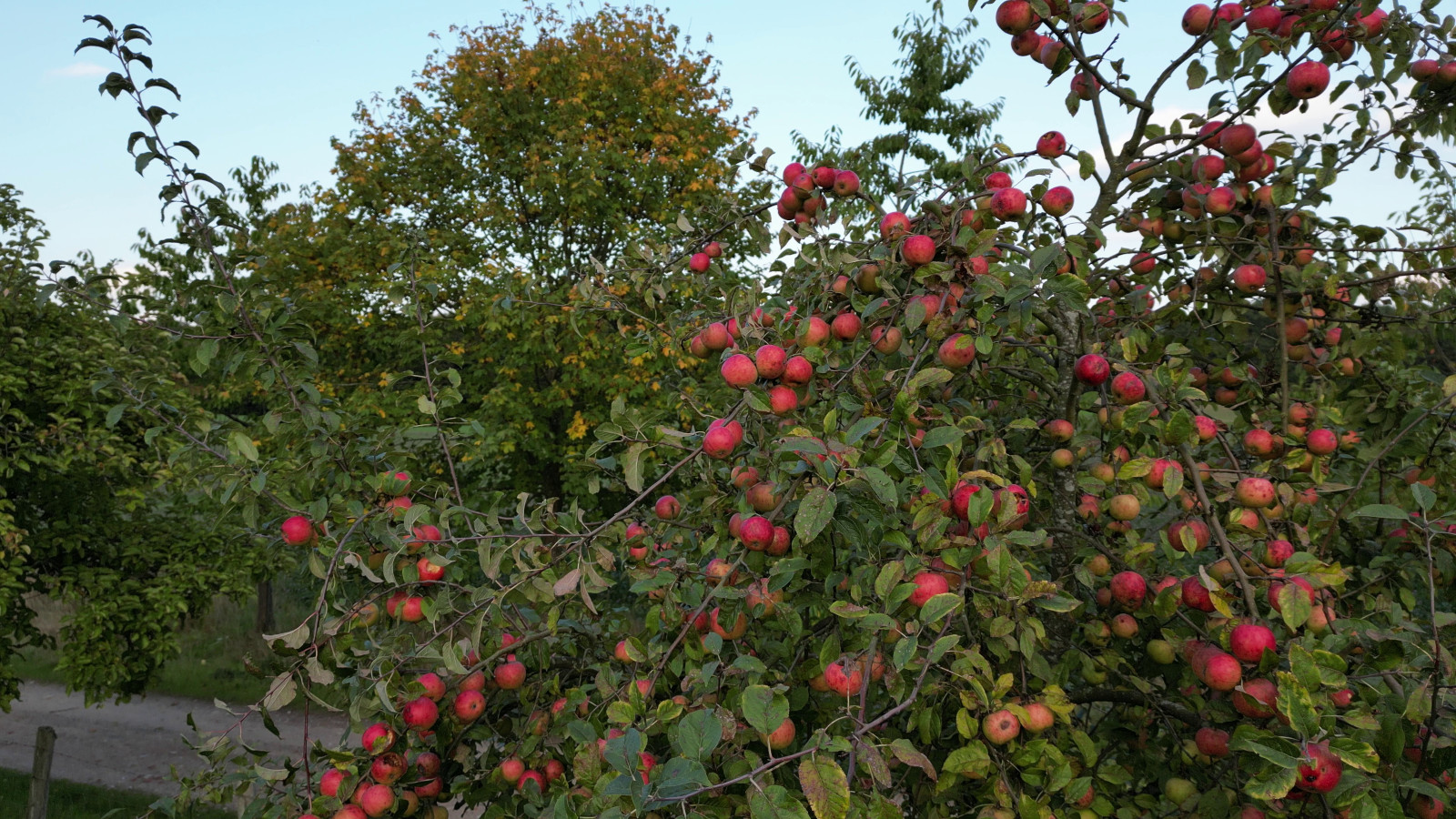} &
\includegraphics[width = .45\linewidth]{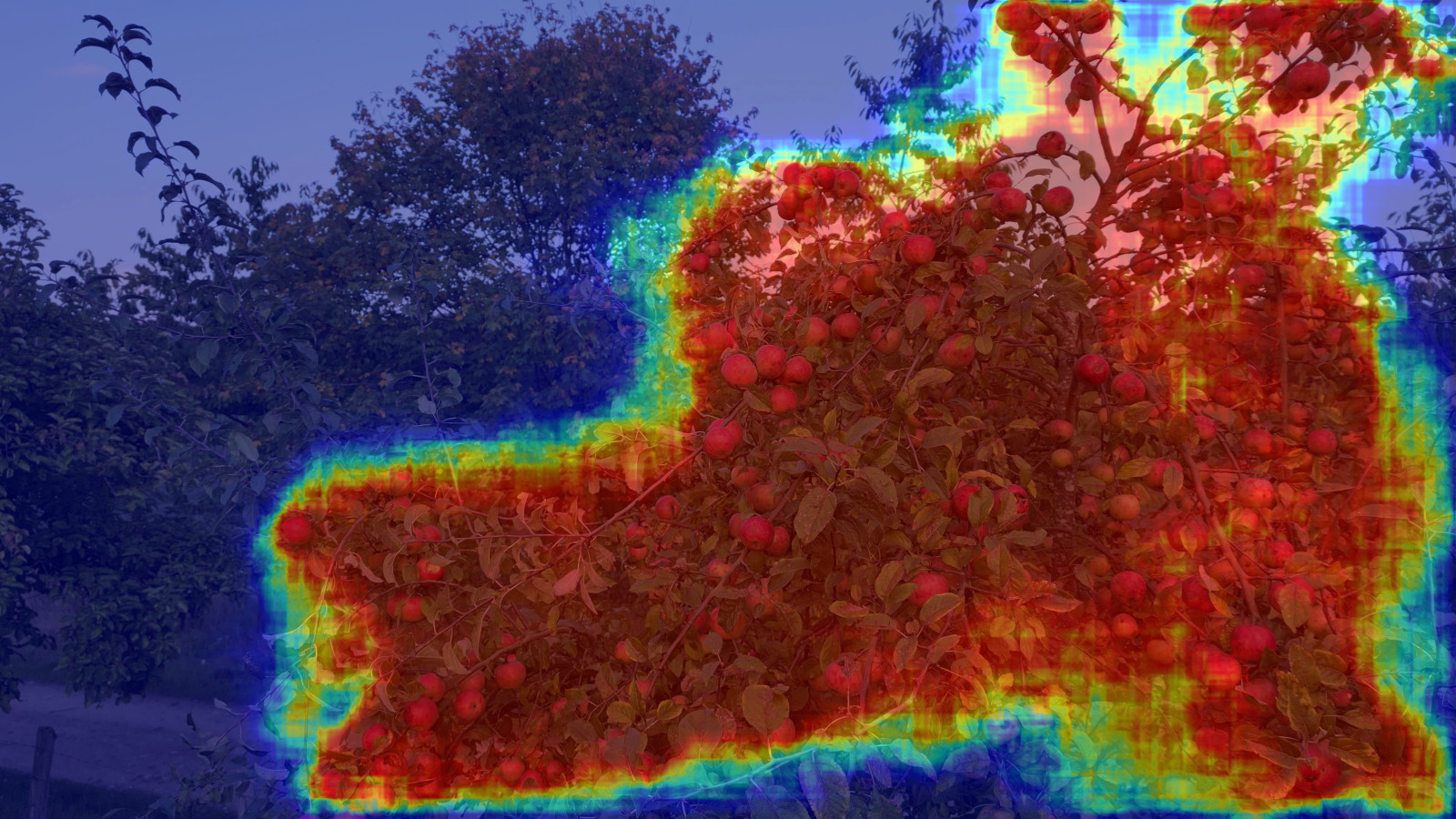}\\
{\footnotesize Input image}&{\footnotesize Attention map}
\end{tabular}
\caption{Attention maps generated by our TreeAttention module that focuses processing on the foreground tree crown. Red areas indicate high attention, while blue areas indicate low attention.
}
\label{fig:TA_qual}
\end{figure}

First, we present the results of our learned TreeAttention, which aims to select all relevant tiles of the foreground tree crown using contextual information. Overall, TreeAttention recalls $99.71\%$ of the labeled apples, while processing only $62.47\%$ of all tiles. The strong recall and the good precision are also well visible from the qualitative results in~\cref{fig:TA_qual} depicting results of TreeAttention encoded as heat maps. While both tree crowns are completely covered (overlaid with red), most of the background is pruned (overlaid with blue). Hence, our TreeAttention is a high-quality pre-processing mechanism to effectively focus the processing, while preserving strong overall detection performance.

\subsubsection{Comparison of Tiling Approaches}
Given the high-quality attention maps generated by TreeAttention, we are able to reduce the number of processed tiles substantially, as shown previously. Now, we compare our selective tiling with the standard tiling that extracts all considered tiles. Table~\ref{tab:tilingComp} shows the result of this comparison for our proposed system in terms of AP and runtime. While the apple detection result in terms of AP is almost constant between the two tiling strategies, the runtime is reduced by $24.2\%$ with our selective tiling approach. Therefore, selective tiling combines the benefits of standard tiling, improving the detection of small objects~(see~\cref{sec:evalProp}), with only a moderate increase in runtime.

\begin{table}[t]
\centering
\caption{Comparison of standard tiling and the proposed selective tiling in S$^3$AD in terms of AP and runtime.}
\label{tab:tilingComp}
\begin{tabular}{lcc}
\toprule
Tiling approach  & AP$\uparrow$ & Runtime[s]$\downarrow$ \\ \midrule  
Standard& 0.423 & 1.98\\
Selective (ours) & 0.423 & 1.51\\
\bottomrule
\end{tabular}
\end{table}

%

\section{Conclusion}
\label{sec:conclusion}
In this paper, we reformulated the problem of apple detection in orchard environments as a semi-supervised detection task to improve the results in the absence of large-scale annotated training data. To this end, we collected MAD, a new apple detection dataset comprising a total of 4,545 high-resolution images and 14,667 annotated apples, including a large share of unlabeled data for semi-supervised training. Taking advantage of MAD, we proposed S$^3$AD, a novel semi-supervised apple detection system based on contextual attention, an efficient selective tiling approach, and a Faster R-CNN detector trained in a Soft Teacher framework. While the tiling approach improves the challenging detection of small apples utilizing the full resolution of the input images, S$^3$AD uses contextual attention to selectively process tiles and limit the additional runtime induced by tiling.

Through extensive evaluation on our novel dataset, which features complex orchard environments, we showed substantial improvements of S$^3$AD compared to strong baselines~(up to $14.9\%$). Further results on the MSU dataset confirmed these improvements. By analyzing the results of S$^3$AD in more detail, we showed that especially on small objects, the selective tiling approach based on contextual information improved the performance, while limiting the additional runtime and maintaining a strong apple detection performance. 
Overall, this paper presents a new reformulation of the apple detection task and a novel, high-quality semi-supervised apple detection approach, addressing our new semi-supervised apple detection task.

\section*{Acknowledgments}
We express our gratitude to Jan-Gerrit Habekost, Silas Ueberschaer, and Jan Willruth for their support in this project. This research was partially funded by the DDLitLab at the University of Hamburg and the Stiftung Innovation in der Hochschullehre. 


{\small
\bibliographystyle{ieee_fullname}
\bibliography{egbib}
}

\end{document}